\pdfoutput=1
\documentclass[10pt]{article}

\usepackage{graphicx}
\usepackage{amsmath}
\usepackage{amssymb}
\usepackage{apacite}
\usepackage{listings}
\usepackage{courier}
\usepackage[vlined,ruled]{algorithm2e}
\newcommand{\ra}{\rangle}
\newcommand{\la}{\langle}

\newcommand{\ttt}{\texttt}

\begin{document}

\title{General-Purpose Computing \\ on a \\ Semantic Network Substrate\footnote{Rodriguez, M.A., ``General-Purpose Computing on a Semantic Network Substrate," Emergent Web Intelligence: Advanced Semantic Technologies, Advanced Information and Knowledge Processing series, eds. R. Chbeir, A. Hassanien,  A. Abraham, and Y. Badr, Springer-Verlag, pages 57-104, ISBN:978-1-84996-076-2, June 2010.}}
\author{Marko A. Rodriguez\\
Digital Library Research and Prototyping Team \\
Los Alamos National Laboratory \\
Los Alamos, New Mexico 87545\\}
\date{original: April 16, 2007 revision: October 7, 2007}
\maketitle{}
\abstract{
This article presents a model of general-purpose computing on a semantic network substrate. The concepts presented are applicable to any semantic network representation. However, due to the standards and technological infrastructure devoted to the Semantic Web effort, this article is presented from this point of view. In the proposed model of computing, the application programming interface, the run-time program, and the state of the computing virtual machine are all represented in the Resource Description Framework (RDF). The implementation of the concepts presented provides a practical computing paradigm that leverages the highly-distributed and standardized representational-layer of the Semantic Web.
\newline \newline
\textbf{keywords:} Resource Description Framework, Web Ontology Language, Virtual Machines, Object-Oriented Programming, Semantic Web Computing
}

\section{Introduction}

This article discusses computing in semantic networks. A semantic network is a directed labeled graph \cite{sowa:semantic1991}. The thesis of this article is that the state of a computing machine, its low-level instructions, and the executing program can be represented as a semantic network. The computational model that is presented can be instantiated using any semantic network representation. However, given the existence of the Resource Description Framework (RDF) \cite{rdfspec:manola2004} and the popular Web Ontology Language (OWL) \cite{owlspec:mcguinness2004}, this article presents the theory and the application in terms of these constructs.

The computing model that is proposed is perhaps simple in theory, but in application, requires a relatively strong background in the computer sciences. This article discusses a wide breadth of concepts including those from computer architecture, the Semantic Web, and object-oriented programming. In order to accommodate all interested readers, each discipline's concepts will be introduced at a tutorial level in this introduction. The remainder of the article presents a more in-depth practical application of the proposed model. The practical application includes a specification for an RDF virtual machine architecture (RVM) called Fhat (pronounced f\u{a}t, like ``fat") and an RDF programming language designed specifically for that architecture called Neno (pronounced n\={e}n\={o}, like ``knee-know").

The introduction to this article is split into three subsections. \S \ref{sec:comparch} provides a brief introduction to the field of computer architecture in order to elucidate those concepts which will be of primary interest later in the article. 
\S \ref{sec:semweb} discusses the Semantic Web and the RDF semantic network data model. Finally, \S \ref{sec:oo} provides an overview of object-oriented programming and its relation to OWL.

\subsection{General-Purpose Computing and the Virtual Machine\label{sec:comparch}}

A general-purpose computer is one that can support any known computation. The meaning of computability and what is required to compute was first developed by Alan Turning in the late 1930s \cite{compute:turing1937}. Since then, and with the large leaps in the engineering of computing machines, most computers of today are general-purpose computing machines. The two primary components of a computing machine are the central processing unit (CPU) and the main memory (RAM). 

The purpose of the CPU is to perform calculations (i.e.~execute algorithms on data). Generally, the CPU reads instructions and data from RAM, performs a calculation, and re-inserts its results back into RAM. Most CPUs maintain a relatively small set of instructions that they can execute \cite{comparch:hennessy2002}. Example instructions include \ttt{add}, \ttt{sub}, \ttt{load}, \ttt{store}, \ttt{branch}, \ttt{goto}, etc. However, these primitive instructions can be composed to perform any computing task desired by the programmer. 

Currently, the smallest unit of information in a digital computer is the bit. A bit can either be a $0$ or a $1$. Bits are combined to form bytes (8-bits) and words (machine architecture dependent). Most desktop machines of everyday use have a 32-bit word and are called 32-bit machines. 32-bits can be used to represent $2^{32}$ different ``things".  For example, an unsigned 32-bit integer can represent the numbers 0 to 4,294,967,295. While instructions, like data, are represented as a series of 0s and 1s, it is possible to represent the instructions in a more human readable form. The abstraction above binary machine language is called assembly language  \cite{mipsassem:britton2003}. For example, the following three assembly instructions
\begin{verbatim}

load 2, 3
load 1, 2
add 1, 2, 3
store 3, 2

\end{verbatim}
instruct the CPU to 1) read the word in the memory cell at memory address $2$ in RAM and store it in CPU register $3$, 2) read the word at memory address $1$ and store it in register $2$, 3) add the contents of register $1$ and $2$ and store the result in register $3$, and finally 4) store the word in register $3$ into memory address $2$ of RAM.

Modern day computer languages are written at a much higher level of abstraction than both machine and assembly language. For instance, the previous instructions could be represented by a single statement as
\begin{verbatim}

z = y + x

\end{verbatim}
where register $1$ holds the value of variable \ttt{y}, register $2$ holds the value of variable \ttt{x} and memory address $3$ holds the value of variable \ttt{z}. 

To the modern-day programmer, the low-level CPU instructions are hidden. It is the role of the language compiler to translate the human readable/writeable source code into the machine code of the CPU. Simply stated, a compiler is a computer program that translates information written in one language to another language \cite{compile:aho1986}. In practice, the compiler translates the human code to a list of CPU instructions that are stored in RAM and executed by the CPU in sequential order as pointed to by the CPU's program counter (PC). In some instances, the compiler will translate the human code to the native language of the CPU (the instruction set of the CPU). In other cases, the compiler will translate the human code to another language for a virtual machine to compute \cite{vm:craig2005}. Virtual machine language is called byte-code. A virtual machine is, for all practical purposes, a CPU represented in software, not hardware. However, depending on the complexity of the implementation, a virtual machine can either do a hard implementation (where an exact software replica of hardware is instantiated) or a soft implementation (where more of the hardware components are black-boxed in software). The virtual machine computes its byte-code instructions by using the underlying hardware CPU's instruction set. Finally, to complete the computation stack, the CPU relies on the underlying physical laws of nature to compute its instructions. The laws of physics drive the CPU from state to state. The evolution of states and its effects on the world is computing.

Perhaps the most popular virtual machine is the Java virtual machine (JVM) \cite{jvm:lindholm1999} of the Java programming language \cite{thinkjava:ecke2002}. The JVM is a piece of software that runs on a physical machine. The JVM has its own instruction set much like a hardware CPU has its own instruction set. The JVM resides in RAM and requires the physical CPU to compute its evolution. Thus, the JVM translates its instructions to the instruction set of the native CPU. The benefit of this model is that irrespective of the underlying hardware CPU, a JVM designed for that CPU architecture can read and process any Java software (i.e.~any Java byte-code). The drawback, is that the computation is slower than when instructions are represented as instructions for the native CPU.  

\subsection{The Semantic Web and RDF\label{sec:semweb}}

The previous section described the most fundamental aspects of computing. This section presents one of the most abstract levels of computing: the Semantic Web. The primary goal of the Semantic Web effort is to provide a standardized framework for describing resources (both physical and conceptual) and their relationships to one another \cite{spinweb:fensel2003}. This framework is called the Resource Description Framework (RDF) \cite{rdfspec:manola2004}.\footnote{Note that RDF is a data model, not a syntax. RDF has many different syntaxes like RDF/XML \cite{rdfspec:manola2004}, Notation 3 (N3) \cite{n3:lee1998}, the N-TRIPLE format \cite{ntriple:beckett2001}, and TRiX \cite{trix:carroll2004}.} 

RDF maintains two central tenets. The first states that the lowest unit of representation is the Universal Resource Identifier (URI) \cite{uri:berners2005} and the literal (i.e.~strings, integers, floating point numbers, etc.).\footnote{There also exist blank or anonymous nodes. There will be no discussion of blank nodes in this article.} A URI unambiguously identifies a resource. When two resources share the same URI, they are the same resource. However, note that when two resources do not share the same URI, this does not necessarily mean that they are not the same resource. In practice, URIs are namespaced to ensure that name conflicts do not occur across different organizations \cite{namespace:bray2006}. For example, the URI \ttt{http://www.newspaper.org/Article} can have a different meaning, or connotation, than \ttt{http://www.science.net/Article} because they are from different namespaces.\footnote{For the sake of brevity, prefixes are usually used instead of the full namespace. For instance, \ttt{http://www.w3.org/1999/02/22-rdf-syntax-ns\#} is prefixed as \ttt{rdf:}.}

\begin{figure}[h!]
	\centering
		\includegraphics[width=0.35\textwidth]{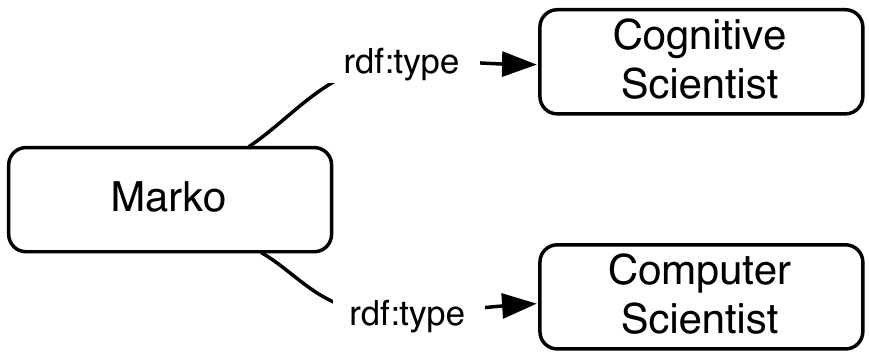}
	\caption{An example RDF network.}
	\label{fig:example-rdf}
\end{figure}

The second tenet of RDF states that URIs and literal values are connected to one another in sets of triples, denoting 
edges of a directed labeled graph. A triple is the smallest relational fact that can be asserted about the world \cite{know:sowa1999}. For instance, the statement ``I am", can be denoted $( \ttt{I}, \ttt{am}, \ttt{I} )$ in triple form. The first element of the triple is called the subject and can be any URI. The second element is called the predicate, and it can also be any URI. Finally, the third element is called the object, and it can be any URI or literal. If $U$ denotes the set of all URIs and $L$ denotes the set of all literals, then an RDF network denoted $G$ can be defined as
\begin{equation*}
G \subseteq (U \times U \times (U \cup L)).
\end{equation*}
RDF has attracted commercial and scholarly interest, not only because of the Semantic Web vision, but because RDF provides a unique way of modeling data. This enthusiasm has sparked the development and distribution of various triple-store applications dedicated to the storage and manipulation of RDF networks \cite{lee:triple2004}. Some triple-stores can support computations on RDF networks that are on the order of $10^{10}$ triples \cite{agraph:aasman2006}. A triple-store is analagous to a relational database. However, instead of representing data in relational tables, a triple-store represents its data as a semantic network. A triple-store provides an interface to an RDF network for the purpose of reading from and writing to the RDF network. The most implemented query language is the SPARQL Protocol and RDF Query Language (SPARQL) \cite{sparql:prud2004}. SPARQL, loosely, is a hybrid of both SQL (a relational database language) and Prolog (a logic programming language) \cite{proglan:louden2003}. As an example, the following SPARQL query returns all URIs that are both a type of \ttt{CognitiveScientist} and \ttt{ComputerScientist}.
 \begin{verbatim}

 SELECT ?x
 WHERE {
   ?x <rdf:type> <ComputerScientist> .
   ?x <rdf:type> <CognitiveScientist> } 
 \end{verbatim}
The example SPARQL query will bind the variable \ttt{?x} to all URIs that are the subject of the triples with a predicate of \ttt{rdf:type} and objects of \ttt{ComputerScientist} and \ttt{CognitiveScientist}.  For the example RDF network diagrammed in Figure \ref{fig:example-rdf}, \ttt{?x} would bind to \ttt{Marko}. Thus, the query above would return \ttt{Marko}.\footnote{Many triple-store applications support reasoning about resources during a query (at run-time). For example, suppose that the triple (\ttt{Marko}, \ttt{rdf:type}, \ttt{ComputerScientist}) does not exist in the RDF network, but instead there exist the triples (\ttt{Marko}, \ttt{rdf:type}, \ttt{ComputerEngineer}) and (\ttt{ComputerEngineer}, \ttt{owl:sameAs}, \ttt{ComputerScientist}). With OWL reasoning, $?x$ would still bind to \ttt{Marko} because  \ttt{ComputerEngineer} and  \ttt{ComputerScientist} are the same according to OWL semantics. The RDF computing concepts presented in this article primarily focus on triple pattern matching and thus, beyond direct URI and literal name matching, no other semantics are used.}

The previous query can be represented in its more set theoretic sense as
\begin{align*}
	X = \; & \{ ?x \; | \; (?x, \ttt{rdf:type}, \ttt{ComputerScientist}) \in G \\
			         & \; \; \; \; \wedge \; (?x, \ttt{rdf:type}, \ttt{CognitiveScientist}) \in G \},
\end{align*}
 where $X$ is the set of URIs that bind to $?x$ and $G$ is the RDF network represented as an edge list. The above syntax's semantics is ``$X$ is the set of all elements $?x$ such that $?x$ is the head of the triple ending with \ttt{rdf:type}, \ttt{ComputerScientist} and the head of the triple ending with \ttt{rdf:type}, \ttt{CognitiveScientist}, where both triples are in the triple list $G$". Only recently has there been a proposal to extend SPARQL to support writing and deleting triples to and from an RDF network. SPARQL/Update \cite{sparqlupdate:seaborne2007} can be used to add the fact that \ttt{Marko} is also an \ttt{rdf:type} of \ttt{Human}.
 \begin{verbatim}
 
INSERT { <Marko> <rdf:type> <Human> . }
 \end{verbatim}
 In a more set theoretic notation, this statement is equivalent to
 \begin{equation*}
	G =  G \cup (\ttt{Marko}, \ttt{rdf:type}, \ttt{Human}).
\end{equation*}
The semantics of the previous statement is: ``Set the triple list $G$ to the current triple list $G$ unioned with the triple (\ttt{Marko}, \ttt{rdf:type}, \ttt{Human}). Finally, it is possible to remove a triple using SPARQL/Update. For instance,
 \begin{verbatim}

DELETE { <I> <am> <I> . }
 \end{verbatim}
In set theoretic notation, this is equivalent to
 \begin{equation*}
	G =  G \setminus (\ttt{I}, \ttt{am}, \ttt{I}),
\end{equation*}
where the semantics are``set the triple list $G$ to the current triple list $G$ minus the triple (\ttt{I}, \ttt{am}, \ttt{I})".

\subsection{Object-Oriented Programming and OWL\label{sec:oo}}

\begin{figure}[h!]
	\centering
		\includegraphics[width=0.25\textwidth]{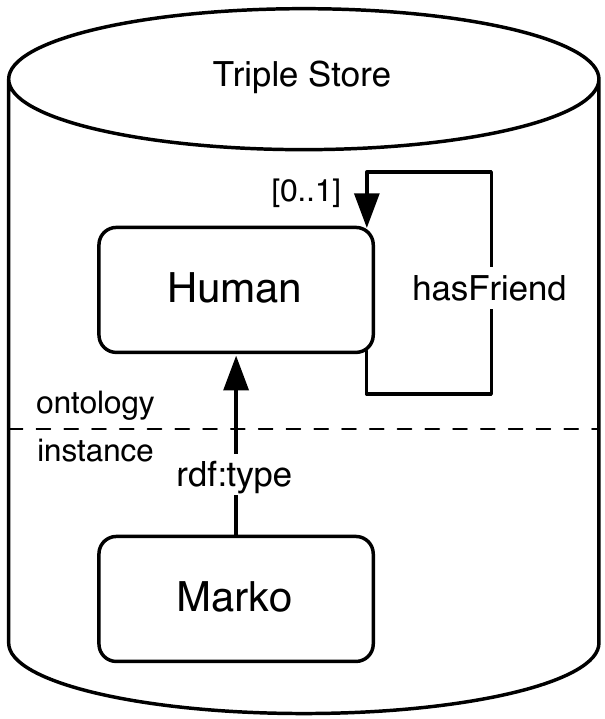}
	\caption{An ontology and an instance is represented in an RDF network.}
	\label{fig:example-friend}
\end{figure}

OWL is an ontology modeling language represented completely in RDF. In OWL, it is possible to model abstract classes and their relationships to one another as well as to use these models and the semantics of OWL to reason about unspecified relationships. In OWL semantics, if \ttt{Human} is a class and there exists the triple \ttt{(Marko, rdf:type, Human}), then \ttt{Marko} is considered an instance of \ttt{Human}. The URI \ttt{Human} is part of the ontology-level of the RDF network and the URI \ttt{Marko} is part of the instance-level (also called the individual-level) of the RDF network. In OWL, it is possible to state that all \ttt{Human}s can have another \ttt{Human} as a friend. This is possible by declaring an \ttt{owl:ObjectProperty} named \ttt{hasFriend} that has an \ttt{rdfs:domain} of \ttt{Human} and an \ttt{rdfs:range} of \ttt{Human}. Furthermore, it is possible to restrict the cardinality of the \ttt{hasFriend} property and thus, state that a \ttt{Human} can have no more than one friend. This is diagrammed in Figure \ref{fig:example-friend}.\footnote{In this article, ontology diagrams will not explicitly represent the constructs \ttt{rdfs:domain}, \ttt{rdfs:range}, nor the \ttt{owl:Restriction} anonymous URIs. These URIs are assumed to be apparent from the diagram. For example, the restriction shown as \ttt{[0..1]} in Figure \ref{fig:example-friend} is represented by an \ttt{owl:Restriction} for the \ttt{hasFriend} property where the \ttt{maxCardinality} is $1$ and \ttt{Human} is an \ttt{rdfs:subClassOf} of this \ttt{owl:Restriction}.}

A class specification in object-oriented programming is called an application programming interface (API) \cite{proglang:sebesta2005}. OWL ontologies share some similarities to the object-oriented API. However, OWL ontologies also differ in many respects. OWL is a description logic language that is primarily focused on a means by which to reason on RDF data. An object-oriented API is primarily focused on concretely defining classes and their explicit relationships to one another and is thus, more in line with the frames modeling paradigm. Furthermore, OWL ontologies can contain instances (i.e.~individuals), allow for multiple inheritance, do not support the unique name assumption, nor the closed world assumption \cite{activerdf:oren2007,frameowl:wang2006}.  

Another aspect of OWL that differs from object-oriented APIs is that object-oriented APIs include the concept of a method. The method is an algorithmic ``behavior" that forms the foundation of the evolutionary processes that drive the instances of these classes from state to state. One of the primary purposes of this article is to introduce an OWL ontology for modeling methods and their low-level machine instructions. While process information can be represented in Frame Logic (i.e.~F-Logic) \cite{flogic:kifer1995}, this article is primarily interested in modeling methods in much the same way that they are represented in modern day object-oriented languages such as Java and C++ and in terms of their syntax, semantics, and low-level representation.

In Java and C++, a method is defined for a class and is used to manipulate the properties (called fields) of an instance of that class. For example,
\begin{verbatim}

class Human {
  Human hasFriend;
  void makeFriend(Human h) {
    this.hasFriend = h;
  }
}
\end{verbatim}
declares that there exists an abstract class called \ttt{Human}. A \ttt{Human} has one field called \ttt{hasFriend}. The \ttt{hasFriend} field refers to an object of type \ttt{Human}. Furthermore, according to the class declaration, a \ttt{Human} has a method called \ttt{makeFriend}. The \ttt{makeFriend} method takes a single argument that is of type \ttt{Human} and sets its \ttt{hasFriend} field to the \ttt{Human} provided in the argument. The \ttt{this} keyword makes explicit that the \ttt{hasFriend} field is the field of the object for which the \ttt{makeFriend} method was invoked.

In many object-oriented languages, an instance of \ttt{Human} is created with the \ttt{new} operator. For instance,
\begin{verbatim}

Human Marko = new Human();

\end{verbatim}
creates a \ttt{Human} named (referenced as) \ttt{Marko}. The \ttt{new} operator is analogous to the \ttt{rdf:type} property. Thus, after this code is executed, a similar situation exists as that which is represented in Figure \ref{fig:example-friend}. However, the ontological model diagrammed in the top half of Figure \ref{fig:example-friend} does not have the \ttt{makeFriend} method URI. The relationship between object-oriented programming and OWL is presented in Table \ref{tab:oordf}.

\begin{table}[h!]
\begin{footnotesize}
\begin{center}
\begin{tabular}{|c|c|c|c|} \hline
& object-oriented & OWL & example \\\hline\hline
class specification & API & ontology & Human \\\hline
object property & field & \ttt{rdf:Property} & hasFriend \\\hline
object method & method &  & makeFriend \\\hline
instantiate & \ttt{new} operator & \ttt{rdf:type} property & \ttt{new}/\ttt{rdf:type} \\\hline
\end{tabular}
\caption{\label{tab:oordf} The relationship between object-oriented programming, OWL, and the section example.}
\end{center}
\end{footnotesize}
\end{table}

It is no large conceptual leap to attach a method URI to a class. Currently, there is no strong  incentive to provide a framework for representing methods in OWL. RDF was originally developed as a data modeling framework, not a programming environment {\it per se}. However, in a similar vein, the Web Ontology Language for Services (OWL-S) has been proposed as a web services model to support the discovery, execution, and tracking of the execution of Semantic Web services \cite{owls:martin2004,devsem:alesso2005}. An OWL-S service exposes a service profile that describes what the service does, a service grounding that describes how to invoke the service, and a service model that describes how the service works. While OWL-S does provide the notion of object-oriented method invocation on the Semantic Web, OWL-S is more at the agent-oriented level and its intended use is for more ``client/server" type problems. Another interesting and related idea is to use RDF as a medium for communication between various computing devices and thus, utilize the Semantic Web as an infrastructure for distributed computing \cite{tripcom:fensel2004}.
 Other object-oriented notions have been proposed within the context of RDF. For instance, SWCLOS \cite{swclos:koide2004} and ActiveRDF \cite{activerdf:oren2007} utilize RDF as a medium for ensuring the long-term persistence of an object. Both frameworks allow their respective languages (CLOS and Ruby) to populate the fields of their objects for use in their language environments. Once their fields have been populated, the object's methods can be invoked in their respective programming environments.

\subsection{The Contributions of this Article}

This article unifies all of the concepts presented hitherto into a framework for computing on RDF networks. In this framework, the state of a computing virtual machine, the API, and the low-level instructions are all represented in RDF. Furthermore, unlike the current programming paradigm, there is no stack of representation. The lowest level of computing and the highest level of computing are represented in the same substrate: URIs, literals, and triples.

This article proposes the concept of OWL APIs, RDF triple-code, and RDF virtual machines (RVM). Human readable/writeable source code is compiled to create an OWL ontology that abstractly represents how instructions should be united to form instruction sequences.\footnote{While OWL has many features that are useful for reasoning about RDF data, the primary purpose of OWL with respect to the concepts presented in this article is to utilize OWL for its ability to create highly restricted data models. These restricted models form the APIs and ensure that instance RDF triple-code can be unambiguously generated by an RVM.} When objects and their methods are instantiated from an OWL API, RDF triple-code is created. RDF triple-code is analogous to virtual machine byte-code, but instead of being represented as bits, bytes, and words, it is represented as URIs and triples. In other words, a piece of executable software is represented as a traversable RDF network. The RVM is a virtual machine whose state is represented in RDF. The RVM's stacks, program counter, frames, etc. are modeled as an RDF network. It is the role of the RVM to ``walk" the traversable RDF triple-code and compute. 

In summary, software is written in human readable/writeable source code, compiled to an OWL API, instantiated to RDF triple-code, and processed by a computing machine whose state is represented in RDF. However, there is always a homunculus. There is always some external process that drives the evolution of the representational substrate. For the JVM, that homunculus is the hardware CPU. For the hardware CPU, the homunculus is the physical laws of nature. For the RVM, the homunculus is some host CPU whether that host CPU is another virtual machine like the JVM or a hardware CPU. Table \ref{tab:levels} presents the different levels of abstraction in computing and how they are represented by the physical machine, virtual machine, and proposed RDF computing paradigms.

\begin{table}[h!]
\begin{footnotesize}
\begin{center}
\begin{tabular}{|c||c|c|c|} \hline
level & machine paradigm & virtual machine paradigm & RDF paradigm \\\hline\hline
high-level code & source code & source code & source code \\\hline
machine code & native instructions & byte-code & triple-code \\\hline
instruction units & bits & bits & URIs and literals \\\hline
machine state & hardware & software & RDF \\\hline
machine execution & physics & hardware & software \\\hline
\end{tabular}
\caption{\label{tab:levels} The various levels of abstraction in current and proposed computing paradigms.}
\end{center}
\end{footnotesize}
\end{table}

\section{A High-Level Perspective}

\begin{figure}[h!]
	\centering
		\includegraphics[width=0.5\textwidth]{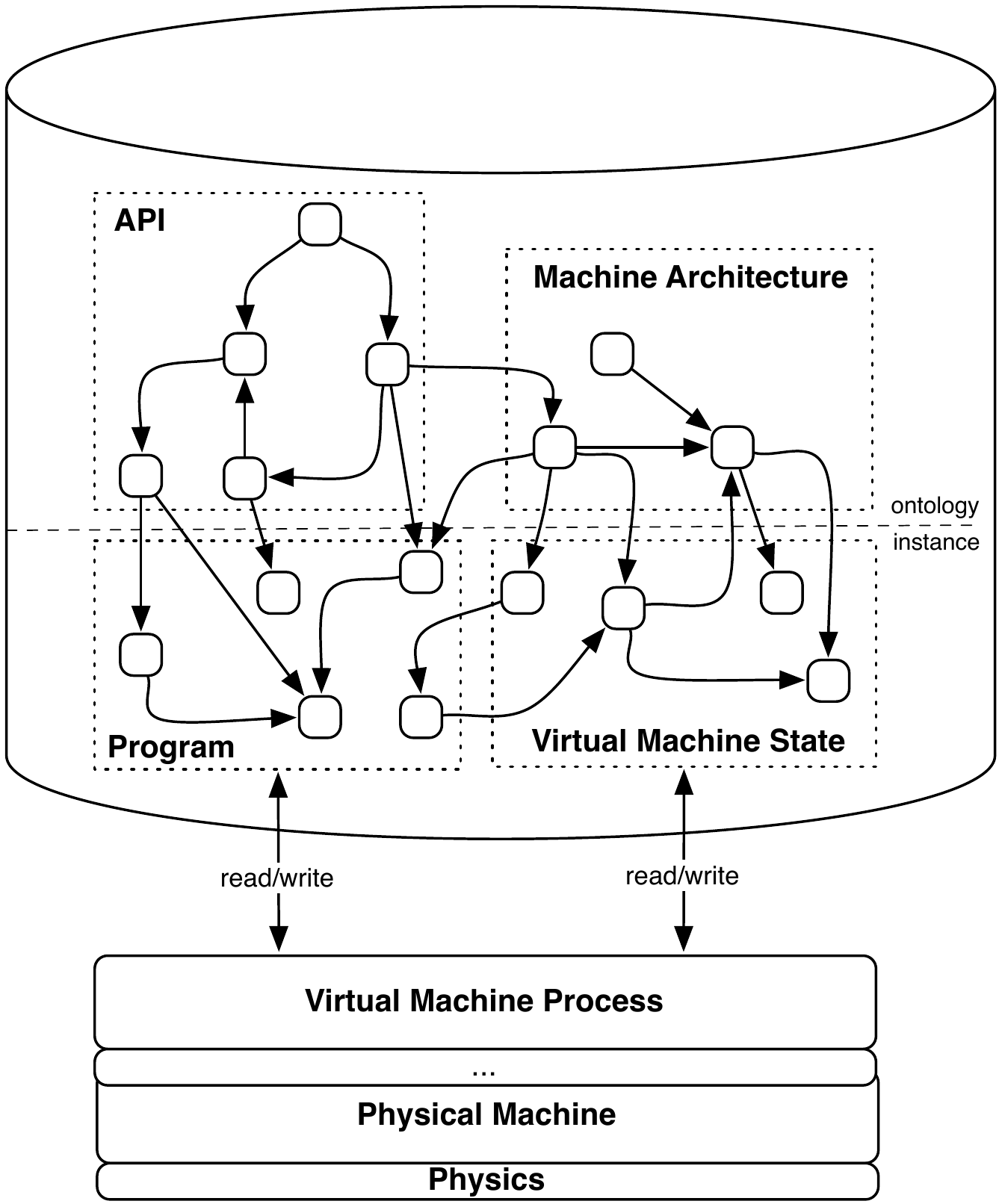}
	\caption{A high-level perspective of the Semantic Web computing environment.}
	\label{fig:system-architecture}
\end{figure}

Assume there exists an RDF triple-store. Internal to that triple-store is an RDF network. That RDF network is composed of triples. A triple is a set of three URIs and/or literals. Those URIs can be used as a pointer to anything. This article presents a model of computation that is represented by URIs and literals and their interrelation to one another (triples). Thus, computation is represented as an RDF network. Figure \ref{fig:system-architecture} presents a high-level perspective on what will be discussed throughout the remainder of this article. What is diagrammed in Figure \ref{fig:system-architecture} is a very compartmentalized model of the components of computing. This model is in line with the common paradigm of computer science and engineering. However, less traditional realizations of this paradigm can remove the discrete levels of representation to support multi-level interactions between the various computing components since all the components are represented in the same RDF substrate: as URIs, literals, and triples.

Figure \ref{fig:system-architecture} shows 6 primary components. Two of these components are at the ontological level of the RDF network, two are at the instance level of the RDF network, and two are at the machine level external to the RDF network. While there are many benefits that emerge from this computing model that are currently seen and as of yet unseen, established interesting aspects are enumerated below.

\begin{enumerate}
\item The total address space of the RVM is the space of all URIs and literals. In the RVM model of computing, the RVM state has no concept of the underlying hardware CPU's address space because instructions and data are represented in RDF. This idea is discussed in \S \ref{sec:uuid}.
\item The Semantic Web is no longer an information gathering infrastructure, but a distributed information processing infrastructure (the process can move to the data, the data doesn't have to move to the process). An RVM can be ``GETed" from a web-server as an RDF/XML document or ``SELECTed" from an RDF triple-store. RDF programs and RVM states are ``first-class" web-entities. The ramifications of this is that an RVM can move between triple-store environments and can compute on local data sets without requiring moving the data to the processor. This idea is discussed in \S \ref{sec:migrate}.
\item This model maintains the ``write once, run anywhere" paradigm of the JVM. The RVM model ensures that human readable/writeable source code is compiled down to an intermediate language that is independent of the underlying hardware CPU executing the RVM process.
\item Languages built on a semantic network substrate can have unique constructs not found in other languages (e.g.~inverse field referencing, multi-instance fields, field querying, etc.). While it is theoretically possible to add these constructs to other languages, they are not provided in the core of the languages as these languages do not have an underlying semantic network data model. These novel language constructs are discussed in \S \ref{sec:lang}.
\item Currently, there already exists an infrastructure to support the paradigm (triple-stores, ontology modeling languages, query languages, etc.) and thus, requires very little investment by the community. The primary investment is the development of source-to-OWL API compilers, RVMs, and the standardization of RDF triple-code and RVM distribution/security protocols.
\item An RVM can be engineered at any level of complexity. It is possible to move the complexity to the software implementing the RVM process to ease machine architecture development and speed up computing time. This idea is discussed in \S \ref{sec:rfhat}.
\item In this model, language reflection exists at the API, software, and RVM level (everything is represented in RDF). This idea is discussed in \S \ref{sec:reflect}.
\end{enumerate}

\subsection{The Ontological Level}

The ontological level of the RDF network diagrammed in Figure \ref{fig:system-architecture} is represented in OWL. This subsection will discuss the two primary ontological components: the API, and the RVM architecture.

\subsubsection{The API}

OWL supports the specification of class interactions. However, class interactions are specified in terms of property relationships, not method invocations. OWL has no formal way of specifying class behaviors (i.e.~methods). However, in OWL, it is possible to define method and instruction classes and formally specify restrictions that dictate how instructions should be interrelated within a method. The method and instruction ontology presented in this article makes RDF a programming framework and not just a data modeling framework.

\subsubsection{The Machine Architecture}

The RDF machine architecture is modeled in OWL. The machine architecture ontology is an abstract description of an instance of a particular RVM. Depending on the level of abstraction required, different machine architectures can be implemented at varying levels of detail.

\subsection{The Instance Level}

The instance level of an RDF network is constrained by the requirements specified in the ontological level of the RDF network. This subsection will present the two components of the instance layer of the diagram in Figure \ref{fig:system-architecture}.

\subsubsection{The Program}

An API abstractly defines a software application. When an API is instantiated, instance RDF triple-code is created. Triple-code represents the instructions used by an RVM to compute.

\subsubsection{The Virtual Machine}

An instance of the machine architecture is an RDF virtual machine (RVM). The purpose of the RVM is to represent its state (stacks, program counter, etc.) in the same RDF network as the triple-code instructions. However, the RDF-based RVM is not a ``true" computer. The RVM simply represents its state in RDF. The RVM requires a software implementation outside the triple-store to compute its instructions. This requires the machine level discussed next.

\subsection{The Machine Level}

The machine level is where the actual computation is executed. An RDF network is a data structure. RDF is not a processor in the common sense---it has no way of evolving itself. In order to process RDF data, some external process must read and write to the RDF network. The reading and writing of the RDF network evolves the RVM and the objects on which it is computing. This section discusses the machine level that is diagrammed in Figure \ref{fig:system-architecture}.

\subsubsection{The Virtual Machine Process}

The virtual machine process is represented in software on a particular host machine. The RVM processor must be compatible with both the triple-store interface (e.g.~SPARQL/Update) and the underlying host machine. The RVM's host machine can be the physical machine (hardware CPU) or another virtual machine. For instance, if the RVM's machine process is implemented in the Java language, then the machine process runs in the JVM. This is diagrammed in Figure \ref{fig:system-architecture} by the \ttt{...} component in between the virtual machine process and the physical machine.

\subsubsection{The Physical Machine}

The physical machine is the actual hardware CPU. The RVM implementation translates the RDF triple-code to the host machine's instruction set. For example, if the RVM process is running on the Intel Core Duo, then it is the role of the RVM process to translate the RDF triple-code to that specified by the Intel Core Duo instruction set. Thus, portability of this architectural model relies on a per host implementation of the RVM. Finally, to complete the computational stack, the laws of physics compute the hardware CPU. Much like the RDF representation of the RVM is a ``snap-shot" representation of a computation, the hardware CPU is a silicon/electron ``snap-shot" representation of a computation.

\section{The Neno Language}

This section presents the specification of a programming language designed to take advantage of a pure RDF computing environment. This language is called Neno. Neno is a high-level object-oriented language that is written in a grammar similar to other object-oriented languages such as Java and C++. However, Neno provides some functionality that is not possible with other languages (i.e.~not explicit in the constructs of other object-oriented languages). This functionality is not due to the sophistication of the Neno language, but instead, is due to the fact that it is written for an RDF substrate and thus, can take advantage of the flexibility of RDF and its read/write interfaces. For this reason, Neno is in a class of languages that is coined semantic network programming languages. The Ripple programming language is another such semantic network programming language \cite{ripple:shinavier2007}. Both Neno and Ripple are Turing complete and thus, can perform any classical (non-quantum) computation.

Neno source code is written in human readable/writeable plain-text like the source code of many other high-level programming languages. Neno source code is compiled by a NenoFhat compiler. The NenoFhat compiler compiles Neno source code to a Fhat OWL API. The Fhat OWL API is analogous to the \ttt{jar} file of Java. A Fhat RVM instantiates (loads) aspects of the API into the instance layer of the RDF network. This instantiated aspect of the API is executable RDF triple-code. A Fhat RVM processes the triple-code and thus, computes. The analogies between the Neno and Java components are presented in Table \ref{tab:nenojava}.

\begin{table}[h!]
\begin{footnotesize}
\begin{center}
\begin{tabular}{|c||c|c|} \hline
artifact & Neno & Java \\\hline\hline
source code & AClass.neno & AClass.java  \\\hline
compiler & nenofhat & javac \\\hline
API & AClass.owl & AClass.class \\\hline
virtual machine & fhat  & java \\\hline
program & RDF network & JVM memory \\\hline
\end{tabular}
\caption{\label{tab:nenojava} The mapping between Neno and Java components.}
\end{center}
\end{footnotesize}
\end{table}

The following examples will only namespace those entities that are not within the namespace \ttt{http://neno.lanl.gov}. Thus, the default namespace is \ttt{http://neno.lanl.gov} (prefixed as \ttt{neno}). The Neno programming language is engineered to be in compliance with OWL and the XML Schema Definition (XSD\index{XSD}) namespaces. OWL provides the concept of classes, inheritance, datatype and class properties, and property restrictions. However, Neno restricts its compiled Fhat OWL APIs to single-parent classes (i.e.~multiple-inheritance is not supported) and holds the closed world assumption (i.e.~only properties that are stated in the ontology can be computed on in a Neno object). This is similar to what is assumed in Java. XSD provides the specification for the literal data types (e.g.~string, integer, float, double, date, time, etc.). The XSD URI namespace prefix is \ttt{xsd}.

The lexicon that will be used to express the following concepts is drawn from object-oriented programming, not OWL. OWL parlance will only be used when completely describing the ``back-end" of a particular aspect of the language. Table \ref{tab:rdfobj} states the relationship between OWL terms and object-oriented programming terms.
\begin{table}[h!]
\begin{footnotesize}
\begin{center}
\begin{tabular}{|c|c|} \hline
OWL & object-oriented languages  \\\hline\hline
\ttt{owl:Class} & Class  \\\hline
\ttt{neno:Method} & Method \\\hline
\ttt{rdf:Property} & Field \\\hline
subject of \ttt{rdf:type} & Object \\\hline
\end{tabular}
\caption{\label{tab:rdfobj} The mapping between the terms in OWL and object-oriented programming.}
\end{center}
\end{footnotesize}
\end{table}

\subsection{The Universally Unique Identifier Address Space\label{sec:uuid}}

Throughout the remainder of this article, Universally Unique Identifiers (UUIDs) will be continually used \cite{uuid:leach2005}. The set of all UUIDs is a subset of the set of all URIs. A UUID is a 128-bit (16-byte) string that can be created in disparate environments with a near zero probability of ever being reproduced. To understand the number of UUIDs that are possible at 128-bits, it would require 1 trillion unique UUIDs to be created every nanosecond for 10 billion years to exhaust the space of all possible UUIDs.\footnote{This fact was taken from Wikipedia at http://en.wikipedia.org/wiki/UUID.} A UUID can be represented as a 36 character hexadecimal string. For example, 6c3f8afe-ec3d-11db-8314-0800200c9a66, is a UUID. The hexadecimal representation will be used in all the following examples. However, for the sake of brevity, since 36 characters is too lengthy for the examples and diagrams, only the first 8 characters will be used. Thus, 6c3f8afe-ec3d-11db-8314-0800200c9a66 will be represented as 6c3f8afe. Furthermore, UUIDs, when used as URIs are namespaced as 
\begin{equation*}
	\ttt{urn:uuid:6c3f8afe-ec3d-11db-8314-0800200c9a66}
\end{equation*}
and for diagrams and examples, is abbreviated as \ttt{urn:uuid:6c3f8afe}.

When Neno source code is compiled to Fhat triple-code, a UUID is created for nearly everything; every instruction class and instruction instance is identified by a UUID. When a Fhat is instantiated, a UUID is created for all the \ttt{rdfs:Resource}s that compose the machine (i.e.~stacks, frames, etc.). In typical programming environments, the programming language and its computing machine are constrained by the size of RAM (and virtual memory with most modern day operating systems). For a 32-bit machine, the maximum size of RAM is approximately 4 gigabytes. This means that there are only $2^{32}$ possible addresses and thus, words in RAM. However, for Neno, no such constraints exist. The space of all UUIDs is the address space of a Fhat RVM (more generally, the space of all URIs and literals is the address space). Fhat does not use RAM for storing its data and instructions, Fhat uses an RDF network. Thus, Fhat does not have any hard constraint on how much memory it ``allocates" for its processing.

\subsection{Class Declarations in Neno Source Code\label{sec:lang}}

Neno source code has a grammar that is very similar to other object-oriented languages. For instance, suppose the following simple class written in the Java programming language:
\begin{footnotesize}
\begin{verbatim}

package gov.lanl.neno.demo;

import java.lang.*;
import java.util.*;

public class Human {
  private String hasName;
  private ArrayList<Human> hasFriend;
  
  public Human (String n) {
    this.hasName = n;
  }

  public void makeFriend(Human h) {
    if(h != this)
      this.hasFriend.add(h);
  }
  
  public void setName(String n) {
    this.hasName = n;
  }
}

\end{verbatim}
\end{footnotesize}
The \ttt{Human} class has two fields named \ttt{hasName} and \ttt{hasFriend}. The field \ttt{hasName} takes a value of \ttt{String} (or \ttt{java.lang.String} to be more specific) and \ttt{hasFriend} takes a value of \ttt{Human}. The \ttt{Human} class has one constructor and one method. A constructor is used to create an object and is a type of method. In Java, a constructor tells the JVM to allocate memory for the object on the heap (i.e.~an object ``pool") and set the object's field values according to the statements in the body of the constructor. The constructor for \ttt{Human} takes a \ttt{String} called \ttt{n} and creates a new \ttt{Human} instance called an object. The \ttt{Human} constructor sets that object's \ttt{hasName} field to \ttt{n}. The \ttt{Human} method is called \ttt{makeFriend}. This method takes a \ttt{Human} with variable name \ttt{h} as an argument. If the object referenced by \ttt{h} is not the \ttt{Human} for which this method was invoked, then the object for which this method was called has \ttt{h} added to its \ttt{hasFriend} field. Note, that unlike the example in Figure \ref{fig:example-friend}, it is possible for a \ttt{Human} object to have multiple friends because of the use of the \ttt{ArrayList<Human>}.\footnote{Java generics as represented by the $<\;>$ notation is supported by Java 1.5+.}

The Neno programming language is similar to Java. The following source code demonstrates how to declare nearly the same class in Neno.\footnote{When there are no ambiguities in naming, the class declaration can be written without prefixes.}
\begin{footnotesize}
\begin{verbatim}

prefix owl: <http://www.w3.org/2002/07/owl>;
prefix xsd: <http://www.w3.org/2001/XMLSchema>;
prefix demo: <http://neno.lanl.gov/demo>;

owl:Thing demo:Human {
  xsd:string hasName[1];
  demo:Human hasFriend[0..*];
  
  !Human(xsd:string n) {
    this.hasName = n;
  }
  
  makeFriend(demo:Human h) {
    if(h != this)
      this.hasFriend =+ h;
  }
  
  setName(xsd:string n) {
    this.hasName = n;
  } 
}
\end{verbatim}
\end{footnotesize}

While the \ttt{Human} class declaration in Java and in Neno are nearly identical, there are a few constructs that make the two languages different. For one, instead of  ``importing" packages, in Neno, namespaces are declared and ontologies are imported.\footnote{Note that languages such as Java and C++ do maintain the concept of package namespaces.} To ease namespace declarations, prefixes are used (e.g.~\ttt{owl}, \ttt{xsd}, and \ttt{demo}). All constructors are denoted by the class name prefixed by the ! symbol. Similarly, though not in the above example, all destructors are denoted by the class name prefixed by the \verb+~+ symbol. Notice that all datatype primitives (e.g.~\ttt{xsd:string}) are from the XSD\index{XSD} namespace. The Fhat RVM is engineered specifically for these datatypes. Perhaps the most unique aspect of the Neno language is the cardinality restriction specifier in the field declaration (e.g.~\ttt{[0..1]}). Because Neno was designed for a semantic network substrate, there is nothing that prevents the same property (i.e.~field) to point to multiple different URIs. In order to demand that there exist no more than one field, the \ttt{[0..1]} notation is used. Note that \ttt{demo:Human} is an \ttt{rdfs:subClassOf} \ttt{owl:Thing} as specified by the \ttt{owl:Thing demo:Human} class description. Class inheritance is specified by the prefix to the declaration of the class name. Note that in Neno, a class can only have a single parent even though OWL supports multiple-inhertiance.\footnote{This constraint does not apply to \ttt{owl:Restriction}s as Neno classes utilize \ttt{owl:Restriction}s to make explicit property restrictions. Thus, excluding \ttt{owl:Restriction} subclassing, a Neno object class can only be the subclass of a single class.} Furthermore, note that all class properties have a universal restriction on the class or datatype value. For example, in the above \ttt{demo:Human} class, the \ttt{hasName} property must have an \ttt{xsd:string} value and all \ttt{hasFriend} properties must have \ttt{demo:Human} values.

In order to demonstrate the relationship between Neno source code and its compiled OWL API, the following simple class example is presented. The class
\begin{footnotesize}
\begin{verbatim}

prefix owl: <http://www.w3.org/2002/07/owl>; 
prefix xsd: <http://www.w3.org/2001/XMLSchema>; 
prefix demo: <http://neno.lanl.gov/demo>; 

owl:Thing demo:Example {
  xsd:integer t[0..1];

  test(xsd:integer n) {
    for(xsd:integer i=0; i < n; i++) {
      this.t = this.t + 1;
    }
  }
}

\end{verbatim}
\end{footnotesize}
has the following OWL RDF/XML representation:
\begin{scriptsize}
\begin{verbatim}
<rdf:RDF
    xmlns:rdf="http://www.w3.org/1999/02/22-rdf-syntax-ns#"
    xmlns:rdfs="http://www.w3.org/2000/01/rdf-schema#"
    xmlns:owl="http://www.w3.org/2002/07/owl#">
  <owl:Ontology rdf:about="http://neno.lanl.gov"/>
  <owl:Ontology rdf:about="http://neno.lanl.gov/demo">
    <owl:imports rdf:resource="http://neno.lanl.gov"/>
  </owl:Ontology>
  ...
  <!-- A PUSHVALUE INSTRUCTION -->

  <owl:Class rdf:about="http://neno.lanl.gov/demo#2271ea72-877c-4090-9f89-...">
    <rdfs:subClassOf rdf:resource="http://neno.lanl.gov#PushValue"/>
    <rdfs:subClassOf>
      <owl:Restriction>
        <owl:onProperty rdf:resource="http://neno.lanl.gov#hasValue"/>
        <owl:allValuesFrom>
          <owl:Class rdf:about="http://neno.lanl.gov/demo#9792cc3c-5600-4660-..."/>
        </owl:allValuesFrom>
      </owl:Restriction>
    </rdfs:subClassOf>
    <rdfs:subClassOf>
      <owl:Restriction>
        <owl:onProperty rdf:resource="http://neno.lanl.gov#nextInst"/>
        <owl:allValuesFrom>
          <owl:Class rdf:about="http://neno.lanl.gov/demo#a80ba54c-5344-4df1-..."/>
        </owl:allValuesFrom>
      </owl:Restriction>
    </rdfs:subClassOf>
  </owl:Class>

  <!-- THE PUSHED VALUE -->

  <owl:Class rdf:about="http://neno.lanl.gov/demo#9792cc3c-5600-4660-bc1f-...">
    <rdfs:subClassOf rdf:resource="http://neno.lanl.gov#LocalDirect"/>
    <rdfs:subClassOf>
      <owl:Restriction>
        <owl:hasValue rdf:datatype="http://www.w3.org/2001/XMLSchema#integer"
        >1</owl:hasValue>
        <owl:onProperty rdf:resource="http://neno.lanl.gov#hasURI"/>
      </owl:Restriction>
    </rdfs:subClassOf>
  </owl:Class>

  <!-- THE NEXT INSTRUCTION AFTER THE PUSHVALUE INSTRUCTION: AN ADD INSTRUCTION -->

  <owl:Class rdf:about="http://neno.lanl.gov/demo#a80ba54c-5344-4df1-91a0-...">
    <rdfs:subClassOf rdf:resource="http://neno.lanl.gov#Add"/>
    <rdfs:subClassOf>
      <owl:Restriction>
        <owl:onProperty rdf:resource="http://neno.lanl.gov#hasLeft"/>
        <owl:allValuesFrom rdf:resource="http://neno.lanl.gov/demo#4c715d16-b6e6-..."/>
      </owl:Restriction>
    </rdfs:subClassOf>
    <rdfs:subClassOf>
      <owl:Restriction>
        <owl:onProperty rdf:resource="http://neno.lanl.gov#hasRight"/>
        <owl:allValuesFrom rdf:resource="http://neno.lanl.gov/demo#fdde7f6f-b9c0-..."/>
      </owl:Restriction>
    </rdfs:subClassOf>
    <rdfs:subClassOf>
      <owl:Restriction>
        <owl:onProperty rdf:resource="http://neno.lanl.gov#nextInst"/>
        <owl:allValuesFrom rdf:resource="http://neno.lanl.gov/demo#e3b8a797-849b-..."/>
      </owl:Restriction>
    </rdfs:subClassOf>
  </owl:Class>
  ...
</rdf:RDF>

\end{verbatim}
\end{scriptsize}
The most important idea to take away from the above Fhat OWL API subset is that the role of the compiler is to generate UUID-named instruction classes that are subclasses of particular Fhat instructions (e.g.~\ttt{PushValue}). These generated instruction classes have \ttt{owl:Restriction}s on them that ensure that instances of these classes are connected to one another in an unambiguous way (e.g.~\ttt{owl:Restriction}s on their respective \ttt{nextInt} property) and that their operand values are made explicit (e.g.~\ttt{owl:Restriction}s on their respective operand properties). This unambiguous instantiation is the RDF triple-code that is created when a Fhat RVM instantiates the API. 

For example, in the above Fhat OWL API snippet, any \ttt{demo:2271ea72} \ttt{PushValue} instruction instance must have one and only one \ttt{hasValue} property. The value of that property must be a \ttt{demo:9792cc3c} \ttt{LocalDirect} value with a \ttt{hasURI} property value of \ttt{"1"$^\wedge$$^\wedge$<xsd:integer>}. The instance of \ttt{demo:2271ea72} must also have a \ttt{nextInst} property that is of \ttt{rdf:type} \ttt{demo:a80ba54c}, where \ttt{demo:a80ba54c} is an \ttt{rdfs:subClassOf} \ttt{Add}. An instance of this \ttt{demo:a80ba54c} \ttt{Add} instruction instructs the Fhat RVM to add its \ttt{hasLeft} operand and its \ttt{hasRight} operands together. This \ttt{demo:a80ba54c} \ttt{Add} also has a \ttt{nextInst} property value that must be an instance of \ttt{demo:e3b8a797}. Though not shown, the \ttt{demo:e3b8a797} is an \ttt{rdfs:subClassOf} \ttt{Set}. In this way, through strict \ttt{owl:Restriction}s, the flow of triple-code can be generated in an unambiguous manner by the Fhat RVM.

The remainder of this section will go over the more salient aspects of the Neno programming language.

\subsubsection{Declaring Namespaces}

Namespaces promote the distributed nature of the Semantic Web by ensuring that there are no URI name conflicts in the ontologies and instances of different organizations \cite{namespace:bray2006}. The Java language has a similar construct called packaging. The package specification in Java supports organizational namespacing. Neno supports the prefixing of namespaces. For example, \ttt{demo:Human} resolves to
\begin{equation*}
\ttt{http://neno.lanl.gov/demo\#Human}.
\end{equation*}

\subsubsection{Datatypes}

Fhat is engineered to handle \ttt{xsd:anySimpleType} and provides specific support for any of its derived types \cite{xsd:biron2004}. The XSD namespace maintains, amongst others: \ttt{xsd:string}, \ttt{xsd:double}, \ttt{xsd:integer}, \ttt{xsd:date}, etc. Example operations include,
\begin{footnotesize}
\begin{verbatim}

"neno"^^xsd:string + "fhat"^^xsd:string
"2007-11-30"^^xsd:date < "2007-12-01"^^xsd:date
"1"^^xsd:integer - "0"^^xsd:integer

\end{verbatim}
\end{footnotesize}
Neno has low-level support for high-level datatype manipulations such as string concatenation, data and time comparisons, date incrementing, etc. Exactly what operations are allowed with what datatypes will be discussed later when describing the Fhat instruction set.

\subsubsection{The \ttt{this} Variable}

The \ttt{this} variable is used in many object-oriented languages to specify the field to be accessed or the method to be invoked. All methods inherently have \ttt{this} as a variable they can use. The same construct exists in Neno with no variation in meaning.

\subsubsection{Field Cardinality}

While Neno is an object-oriented language, it is also a semantic network programming language. Neno is more in line with the concepts of RDF than it is with those of Java and C++. One of the major distinguishing features of an object in Neno is that objects can have multi-instance fields. This means that a single field (predicate) can have more than one value (object). For instance, in Java
\begin{footnotesize}
\begin{verbatim}

Human marko = new Human("Marko Rodriguez");
marko.setName("Marko Antonio Rodriguez");

\end{verbatim}
\end{footnotesize}
will initially set the \ttt{hasName} field of the \ttt{Human} object referenced by the variable name \ttt{marko} to ``Marko Rodriguez". The invocation of the \ttt{setName} method of \ttt{marko} will replace ``Marko Rodriguez" with ``Marko Antonio Rodriguez". Thus, the field \ttt{hasName} has a cardinality of 1. All fields in Java have a cardinality of 1 and are universally quantified for the specified class (though taxonomical subsumption is supported).

In Neno, it is possible for a field to have a cardinality greater than one. In Neno, when a class' fields are declared, the cardinality specifier is used to denote how many properties of this type are allowed for an instance of this class. Thus, in the Neno code at the start of this section, 
\begin{footnotesize}
\begin{verbatim}

 xsd:string hasName[1];

\end{verbatim}
\end{footnotesize}
states that any \ttt{Human} object must have one and only one field (property) called \ttt{hasName} and that \ttt{hasName} field points to some \ttt{xsd:string}. Therefore, it is illegal for the Fhat RVM to add a new \ttt{hasName} property to the class \ttt{marko}. The original property must be removed before the new property can be added. The general grammar for field restrictions in Neno is \ttt{[\# (..(\# | *))]}, where \ttt{\#} refers to some integer value.

Neno does not adopt any of the OWL semantics regarding cardinality and ``semantically distinct" resources. The \ttt{owl:sameAs} relationship between resources is not considered when determining the cardinality of a property and thus, only the explicit number of properties (explicit triples) of a particular type (predicate) are acknowledged by the NenoFhat compiler and Fhat RVM.

\subsubsection{Handling Fields}

Neno provides the following field and local variable operators: \ttt{=+}, \ttt{=-}, \ttt{=/}, and \ttt{=}. These operators are called ``set plus", ``set minus", ``set clear", and ``set", respectively. The definition of these operators is made apparent through examples that demonstrate their use. For instance, from the class declarations above, the \ttt{Human} class has the field \ttt{hasFriend}. For the Java example, the \ttt{hasFriend} field can have more than one \ttt{Human} value only indirectly through the use of the \ttt{ArrayList<Human>} class. In Neno, no \ttt{ArrayList<Human>} is needed because a field can have a cardinality greater than 1. The cardinality specifier \ttt{[0..*]} states that there are no restrictions on the number of friends a \ttt{Human} can have. In order to add more friends to a \ttt{Human} object, the \ttt{=+} operator is used. If the \ttt{Human} instance has the URI \ttt{urn:uuid:2db4a1d2} and the provided \ttt{Human} argument has the URI \ttt{urn:uuid:47878dcc} then the \ttt{=+} operator instructs Fhat to execute
\begin{footnotesize}
\begin{verbatim}

INSERT { <urn:uuid:2db4a1d2> <demo:hasFriend> <urn:uuid:47878dcc> .}

\end{verbatim}
\end{footnotesize}
on the triple-store. On the other hand, if the \ttt{=} operator was used, then Fhat would issue the following commands to the triple-store:
\begin{footnotesize}
\begin{verbatim}

DELETE { <urn:uuid:2db4a1d2> <demo:hasFriend> ?x .}
INSERT { <urn:uuid:2db4a1d2> <demo:hasFriend> <urn:uuid:47878dcc> .}

\end{verbatim}
\end{footnotesize}
For a multi-instance field, the $=$ is a very destructive operator. For a \ttt{[0..1]} or \ttt{[1]} field, \ttt{=} behaves as one would expect in any other object-oriented language. Furthermore, for a \ttt{[0..1]} or \ttt{[1]} field, \ttt{=+} is not allowed as it will cause the insertion of more than one property of the same predicate.

In order to control the removal of fields from a multi-instance field, the \ttt{=-} and \ttt{=/} operators can be used. For example, suppose the following method declaration in Neno
\begin{footnotesize}
\begin{verbatim}

makeEnemy(Human h) {
  this.hasFriends =- h;
}

\end{verbatim}
\end{footnotesize}
The \ttt{makeEnemy} method will remove the \ttt{Human} object identified by the variable name \ttt{h} from the \ttt{hasFriend} fields. If the \ttt{h} variable is a reference to the URI \ttt{urn:uuid:4800e2c2}, then at the Fhat level, Fhat will execute the following command on the triple-store:
\begin{footnotesize}
\begin{verbatim}

DELETE { <urn:uuid:2db4a1d2> <demo:hasFriend> <urn:uuid:4800e2c2> .}

\end{verbatim}
\end{footnotesize}

Finally, assume that there is a rogue \ttt{Human} that wishes to have no friends at all. In order for this one man army to sever his ties, the $=/$ operator is used. Assume the following overloaded method declaration for a \ttt{Human}.
\begin{footnotesize}
\begin{verbatim}

makeEnemy() {
  this.hasFriends =/;
}

\end{verbatim}
\end{footnotesize}
The above statement statement would have Fhat execute the following delete command on the triple-store:
\begin{scriptsize}
\begin{footnotesize}
\begin{verbatim}

DELETE { <urn:uuid:2db4a1d2> <demo:hasFriend> ?human }

\end{verbatim}
\end{footnotesize}
\end{scriptsize}

\subsubsection{Field Querying}

In many cases, a field (i.e.~property) will have many instances. In computer programming terms, fields can be thought of as arrays. However, these ``arrays" are not objects, but simply greater than one cardinality fields. In Java, arrays are objects and high-level array objects like the \ttt{java.util.ArrayList} provide functions to search an array. In Neno, there are no methods that support such behaviors since fields are not objects. Instead, Neno provides language constructs that support field querying. For example, suppose the following method
\begin{footnotesize}
\begin{verbatim}

boolean isFriend(Human unknown) {
  if(this.hasFriend =? unknown) {
    return true;
  }
  else {
   return false;
  }
 }
\end{verbatim}
\end{footnotesize}
In the above \ttt{isFriend} method, the provided \ttt{Human} argument referenced by the variable name \ttt{unknown} is checked against all the \ttt{hasFriend} fields. Again, the \ttt{owl:sameAs} property is not respected and thus, ``sameness" is determined by exact URIs. The \ttt{=?} operator is a conditional operator and thus, always returns either \ttt{"true"$^\wedge$$^\wedge$xsd:boolean} or \ttt{"false"$^\wedge$$^\wedge$xsd:boolean}. At the Fhat level, if \ttt{this} references the UUID \ttt{urn:uuid:2d386232} and \ttt{unknown} references \ttt{urn:uuid:75e05c12}, then the Fhat RVM executes the following query on the triple-store:
\begin{footnotesize}
\begin{verbatim}

ASK { <urn:uuid:2d386232> <demo:hasFriend>  <urn:uuid:75e05c12> . }

\end{verbatim}
\end{footnotesize}

Similarly, imagine the following method,
\begin{footnotesize}
\begin{verbatim}

boolean isFriendByName(Human unknown) {
  if(this.hasFriend.hasName =? unknown.hasName) {
    return true;
  }
  else {
   return false;
  }
 }
\end{verbatim}
\end{footnotesize}
Assuming the same UUID references for \ttt{this} and \ttt{unknown} from previous examples, the \ttt{=?} operation would have the Fhat execute the following query on the RDF network
\begin{footnotesize}
\begin{verbatim}

ASK { <urn:uuid:2d386232> <demo:hasFriend>  ?x .
      ?x <demo:hasName> ?y .
      <urn:uuid:75e05c12> <demo:hasName> ?y }

\end{verbatim}
\end{footnotesize}

Again, there is no reasoning involved in any of these triple-store operations; only ``raw" triple and URI/literal matching is used.

\subsubsection{Looping and Conditionals}

Looping and conditionals are nearly identical to the Java language. In Neno, there exists the \ttt{for}, \ttt{while}, and \ttt{if/else} constructs. For example, a \ttt{for} statement is
\begin{footnotesize}
\begin{verbatim}

for(xsd:integer i = "0"^^xsd:integer; i<"10"^^xsd:integer; i++) 
{ /* for block */ }

\end{verbatim}
\end{footnotesize}
a \ttt{while} statement is
\begin{footnotesize}
\begin{verbatim}

while(xsd:integer i < "10"^^xsd:integer) 
  { /* while block */ }

\end{verbatim}
\end{footnotesize}
and an \ttt{if/else} statement is
\begin{footnotesize}
\begin{verbatim}

if(xsd:integer i < "10"^^xsd:integer) 
  { /* if block */ } 
  { /* else block */}

\end{verbatim}
\end{footnotesize}

It is important to note that these statements need not have the literal type specifier (e.g. \ttt{xsd:integer}) on every hardcoded literal. The literal type can be inferred from its context and thus, is automatically added by the compiler. For example, since \ttt{i} is an \ttt{xsd:integer}, it is assumed that $10$ is also.

\subsubsection{Field Looping}

In many cases it is desirable to loop through all the resources of a field for the purposes of searching or for manipulating each resource. For instance, suppose the following \ttt{Human} method:
\begin{footnotesize}
\begin{verbatim}

namelessFaces() {
  for(Human h : this.hasFriend) {
    h.hasName = "..."^^xsd:string;
  }
  for(xsd:integer i=0; i<this.hasFriend*; i++) {
    Human h = this.hasFriend[i];
    h.hasName = "."^^xsd:string;
  }
 }
 
\end{verbatim}
\end{footnotesize}
The above \ttt{namelessFaces} method demonstrates two types of field looping mechanisms offered by Neno. The first is analogous to the Java 1.5 language specification. With the first for loop, the variable \ttt{h} is set to a single \ttt{hasFriend} of \ttt{this}. The second for loop uses the index \ttt{i} that goes from index 0 to the size of the ``array" (\ttt{this.hasFriend*}). The \ttt{*} notation in this context returns the number of \ttt{hasFriend} properties of the \ttt{this} object. In other words \ttt{*} returns the cardinality of the \ttt{this.hasFriend} field. 

Finally, as field values are not stored in a vector, but instead as an unordered set, the field ``arrays" in Neno are not guaranteed to be ordered. Thus, \ttt{this.hasFriend[1]} may not be the same value later in the code. Ordering is dependent upon the triple-store's indexing algorithm and stability of a particular order is dependent upon how often re-indexing occurs in the triple-store. It is worth noting that higher-order classes can be created such as specialized \ttt{rdf:Seq} and \ttt{rdf:List} classes to provided ordered support for arrays.

\subsubsection{Type Checking}

The \ttt{typeof} operator can be used to determine the class type of a URI. For instance, the following statement,
\begin{footnotesize}
\begin{verbatim}

xsd:boolean isType = urn:uuid:2db4a1d2 typeof Human
 
\end{verbatim}
\end{footnotesize}
would return \ttt{true} if \ttt{urn:uuid:2db4a1d2} is \ttt{rdf:type} \ttt{Human} or \ttt{rdf:type} of some class that is an \ttt{rdfs:subClassOf} \ttt{Human}. Also,
\begin{footnotesize}
\begin{verbatim}

xsd:boolean isType = urn:uuid:2db4a1d2 typeof rdfs:Resource
 
\end{verbatim}
\end{footnotesize}
always returns \ttt{true}. Thus, RDFS subsumption semantics are respected and thus, Neno respects the subclassing semantics employed by modern objected-oriented languages. Similarly the \ttt{typeOf?} operator returns the type of the resource. For instance,
\begin{footnotesize}
\begin{verbatim}

xsd:anyURI type = urn:uuid:2db4a1d2 typeof?
 
\end{verbatim}
\end{footnotesize}
returns \ttt{http://neno.lanl.gov/demo\#Human}.

\subsubsection{Inverse Field Referencing}

In object-oriented languages the ``dot" operator is used to access a method or field of an object. For instance, in \ttt{this.hasName}, on the left of the ``dot" is the object and on the right of the ``dot" is the field. Whether the right hand side of the operator is a field or method can be deduced by the compiler from its context. If \ttt{this} resolves to the URI \ttt{urn:uuid:2db4a1d2}, then the following Neno code
\begin{footnotesize}
\begin{verbatim}

Human h[0..*] = this.hasFriend;

\end{verbatim}
\end{footnotesize}
would instruct Fhat to execute the following query:
\begin{footnotesize}
\begin{verbatim}

SELECT ?h
  WHERE { <urn:uuid:2db4a1d2> <demo:hasFriend> ?h . }

\end{verbatim}
\end{footnotesize}
According to the previous query, everything that binds to \ttt{?h} will be set to the variable \ttt{h}. The above query says ``locate all \ttt{Human} \ttt{hasFriend}s of \ttt{this} object." However, Neno provides another concept not found in other object-oriented languages called the ``dot dot" operator. The ``dot dot" operator provides support for what is called inverse field referencing (and inverse method invocation discussed next). Assume the following line in some method of some class,
\begin{footnotesize}
\begin{verbatim}

Human h[0..*] = this..hasFriend;

\end{verbatim}
\end{footnotesize}
The above statement says, ``locate all \ttt{Human}s that have \ttt{this} object as their \ttt{hasFriend}." At the Fhat level, Fhat executes the following query on the triple-store:
\begin{footnotesize}
\begin{verbatim}

SELECT ?h
  WHERE { ?h <demo:hasFriend> <urn:uuid:2db4a1d2> .}

\end{verbatim}
\end{footnotesize}
Furthermore, if the statement is 
\begin{footnotesize}
\begin{verbatim}

Human h[0..3] = this..hasFriend;

\end{verbatim}
\end{footnotesize}
Fhat would execute:
\begin{footnotesize}
\begin{verbatim}

SELECT ?h 
  WHERE { ?h <demo:hasFriend> <urn:uuid:2db4a1d2> .} LIMIT 3

\end{verbatim}
\end{footnotesize}

\subsubsection{Inverse Method Invocation}

Like inverse field referencing, inverse method invocation is supported by Neno. Inverse method invocation will invoke all the methods that meet a particular requirement. For instance,
\begin{footnotesize}
\begin{verbatim}

this..hasFriend.makeEnemy(this);

\end{verbatim}
\end{footnotesize}
will ensure that all objects that have \ttt{this} as their friend are no longer friends with \ttt{this}.

\subsubsection{Variable Scoping}

Variable scoping in Neno is equivalent to Java. For example, in
\begin{footnotesize}
\begin{verbatim}

xsd:integer a = "11"^^xsd:integer;
if(a < "10"^^xsd:integer) {
  xsd:integer b = "2"^^xsd:integer;
}
else {
  xsd:integer c = "3"^^xsd:integer;
}

\end{verbatim}
\end{footnotesize}
the true and false block of the \ttt{if} statement can read the variable \ttt{a}, but the true block can not read the \ttt{c} in the false block and the false block can not read the \ttt{b} in the true block. Also, methods are out of scope from one another. The only way methods communicate are through parameter passing, return values, and object manipulations.

\subsubsection{Constructors and Destructors}

Constructors and destructors are used in object-oriented languages to create and destroy object, respectively. The concept of a constructor in Neno is similar to that of Java and C++. The concept of a destructor does not exist in Java, but does in C++. It is very important in Neno to provide the programmer an explicit way of performing object destruction. Again, unlike Java, Neno is intended to be used on a persistent semantic network substrate. Thus, when a Fhat stops executing or an object is no longer accessible by a Fhat, that object should not be automatically removed. In short, Fhat does not provide automatic garbage collection \cite{proglan:louden2003}. It is the role of the programmer to explicitly remove all unwanted objects from the RDF network.

\begin{figure}[h!]
	\centering
		\includegraphics[width=0.75\textwidth]{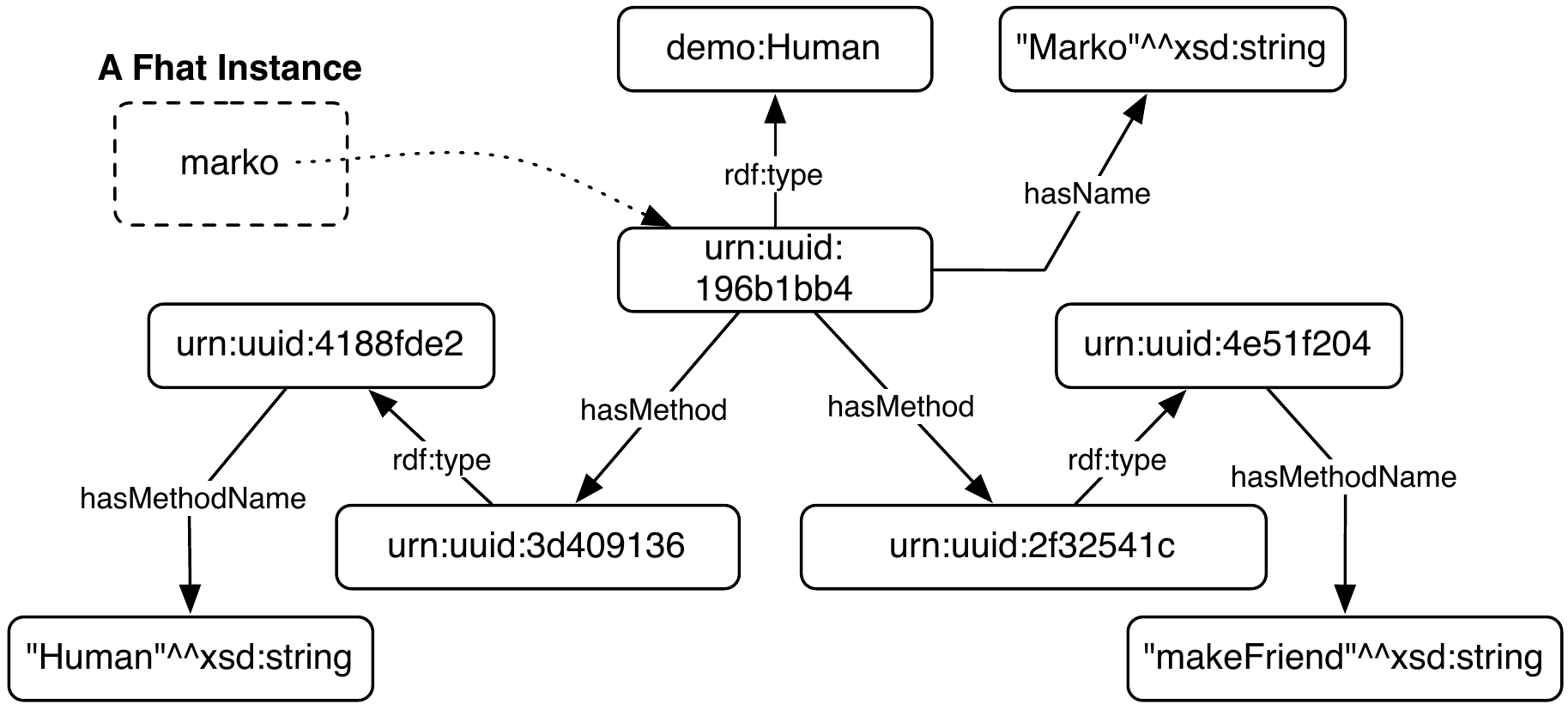}
	\caption{A Fhat instance maintains a variable reference to an object.}
	\label{fig:example-human-instance}
\end{figure}

In order to create a new object, the constructor of a class is called using the \ttt{new} operator. For example,
\begin{footnotesize}
\begin{verbatim}

Human marko = new Human("Marko"^^xsd:string);

\end{verbatim}
\end{footnotesize}
will generate a sub-network in the RDF network equivalent to Figure \ref{fig:example-human-instance}.

The algorithm by which Fhat creates the RDF sub-network will be discussed in the next section. For now, understand that in the variable environment of a Fhat instance there exists a variable named \ttt{marko} that points to the newly created \ttt{Human} instance (e.g.~$\la \ttt{marko}, \ttt{rdf:type}, \ttt{Human} \ra$).

A destructor will instruct Fhat to destroy an object. A destructor is specified in the class declaration. For instance, suppose the following specification for \ttt{demo:Human}:
\begin{footnotesize}
\begin{verbatim}

Thing Human {
  string hasName[1];
  Human hasFriend[0..*];
  
  !Human(string n) {
    this.hasName = n;
  }
  
  ~Human() {
    this.hasName =/
    this.hasFriend =/
    this..hasFriend =/
  } 
}

\end{verbatim}
\end{footnotesize}
In the above class declaration \ttt{!Human(string n)} is a constructor and \verb+~+\ttt{Human()} is a destructor. A destructor is called using the \ttt{delete} operator. For instance,
\begin{footnotesize}
\begin{verbatim}

delete marko;

\end{verbatim}
\end{footnotesize}
calls \ttt{marko}'s \verb+~+\ttt{Human()} destructor.

A class can only have at most one destructor and the destructor takes no arguments. The \verb+~+\ttt{Human()} destructor removes the reference to the object's name, removes all the references to the object's friends, and removes all \ttt{hasFriend} references to that object. Thus, if the \ttt{Human} object has the URI \ttt{urn:uuid:55b2a3b0}, Fhat would execute the following commands on the triple-store:
\begin{footnotesize}
\begin{verbatim}

DELETE { <urn:uuid:55b2a3b0> <demo:hasName> ?name .}
DELETE { <urn:uuid:55b2a3b0> <demo:hasFriend> ?human .}
DELETE { ?human <demo:hasFriend> <urn:uuid:55b2a3b0> .}

\end{verbatim}
\end{footnotesize}
Behind the scenes, Fhat would also remove all the method references of \ttt{urn:uuid:55b2a3b0}, internal variable references to \ttt{urn:uuid:55b2a3b0}, and the \ttt{rdf:type} relationships that relate the object to the ontological-layer. When an object is properly destroyed, only its instance is removed from the RDF network. The object's class specification still exists in the ontological-layer.

\subsubsection{General Query}

In many instances, Fhat will not have a reference to a particular object. Again, the environment anticipated is one in which objects persist in the RDF network. Thus, when code is executed, it is necessary to locate the URI of a particular object for processing. In order to make this easy for the programmer, a query operator is defined called the ``network query" operator and is denoted by the symbol \ttt{<?}. For example,
\begin{footnotesize}
\begin{verbatim}

xsd:string x = "Marko Antonio Rodriguez"^^xsd:string;
xsd:string query = 
   "SELECT ?x WHERE { ?x <demo:hasName> <" + x + "> } 
     LIMIT 1"^^xsd:string;
Human h[0..1] <? query;

\end{verbatim}
\end{footnotesize}
will query the RDF network for at most one \ttt{Human} named ``Marko Antonio Rodriguez". Note that three statements above could have been written as one. However, to demonstrate string concatenation and variable use, three were used.

\subsection{Starting a Program in Neno}

In Neno, there are no static methods. Thus, there does not exist something like the \ttt{public static void main(String[] args)} method in Java. Instead, Fhat is provided a class URI and a method for that class that takes no arguments. The class is automatically instantiated by Fhat and the specified no-argument method is invoked. For example, if Fhat is pointed to the following \ttt{Test} class and \ttt{main} method, then the \ttt{main} method creates a \ttt{Human}, changes its name, then exits. When \ttt{main} exits, Fhat halts.
\begin{footnotesize}
\begin{verbatim}

owl:Thing demo:Test {
  main() {
    demo:Human h = new Human("Marko Rodriguez");
    h.setName("Marko Antonio Rodriguez");
  }
}

\end{verbatim}
\end{footnotesize}

\subsection{Typical Use Case}

This section describes how a developer would typically use the Neno/Fhat environment. The terminal commands below ensure that the NenoFhat compiler translates Neno source code to a Fhat OWL API, loads the Fhat OWL API into the triple-store, instantiates a Fhat RVM, and points the RVM to the \ttt{demo:Test} class with a \ttt{main} method. Note that the third command is broken into four lines for display purposes. Do not assume that there is a newline character at the end of the first three lines of the third statement.
\begin{footnotesize}
\begin{verbatim}

> nenofhat Human.neno -o ntriple -t http://www.triplestore.net/sparql
> nenofhat Test.neno -o xml -t http://www.triplestore.net/sparql
> fhat -vmc http://neno.lanl.gov/neno#Fhat 
       -c http://neno.lanl.gov/neno/demo#Test
       -cm main
       -t http://www.triplestore.net/sparql

\end{verbatim}
\end{footnotesize}
The first terminal command compiles the \ttt{Human.neno} source code into a Fhat OWL API represented in N-TRIPLE format and then inserts the \ttt{Human.ntriple} triples into the triple-store pointed to by the ``-t" URL. The second terminal command compiles the \ttt{Test.neno} source code and generates a Fhat OWL API in RDF/XML called \ttt{Test.xml}. That RDF/XML file is then loaded into the triple-store. The \ttt{nenofhat} compiler can produce any of the popular RDF syntaxes. While in most cases, one or another is chosen, two different syntaxes are shown to demonstrate what is possible with the compiler. Finally, a \ttt{Fhat} processor is initiated. The virtual machine process (\ttt{fhat}) is called with a pointer to an ontological model of the desired machine architecture. The machine architecture is instantiated. The instantiated \ttt{Fhat} then instantiates a \ttt{Test} object and calls its \ttt{main} method. The instantiated \ttt{Test} \ttt{main} method is executable RDF triple-code.

In some instances, a Fhat RVM state may already exist in the triple-store. In such cases, the following command can be invoked to point the Fhat RVM process to the stored RVM state. In the example below, assume that \ttt{urn:uuid:60ab17c2} is of \ttt{rdf:type} \ttt{Fhat}.
\begin{footnotesize}
\begin{verbatim}

> fhat -vmi urn:uuid:60ab17c2 -t http://www.triplestore.net/sparql

\end{verbatim}
\end{footnotesize}
When the \ttt{Fhat} RVM state is located, \ttt{fhat} processes the current instruction pointed to by its \ttt{programLocation}. 

The following list outlines the flags for the \ttt{nenofhat} compiler,

\begin{itemize}
\item -o : output type (ntriple $|$ n3 $|$ xml)
\item -t : triple-store interface
\end{itemize}

and the \ttt{fhat} RVM process,

\begin{itemize}
\item -vmi : virtual machine instance URI
\item -vmc : virtual machine class URI
\item -c : start class URI
\item -cm : start class no-argument method
\item -t : triple-store interface.
\end{itemize}

\section{The Fhat Virtual Machine Architecture\index{RVM!Fhat}}

Fhat is an RVM that was specifically designed for RDF-based semantic network languages. Fhat is a semi-hard implementation of a computing machine. Table \ref{tab:softhard} presents an explanation of the various levels of virtual machine implementations. The concept of soft, semi-hard, and hard implementations are developed here and thus, are not part of the common lexicon. In the JVM, all of the ``hardware" components are represented in software and the state of the machine is not saved outside the current run-time environment. For VHSIC Hardware Description Language (VHDL) machines, the hardware components are modeled at the level of logic gates (AND, OR, XOR, NOT, etc.) \cite{vhdl:coelho1988}. In Fhat, the hardware components are modeled in RDF (the state), but component execution is modeled in software (the process). 

\begin{table}[h!]
\begin{footnotesize}
\begin{center}
\begin{tabular}{|c|c|c|} \hline
implementation type & requirements & example  \\\hline\hline
soft & hardware methods & Java Virtual Machine, r-Fhat  \\\hline
semi-hard & high-level components & Fhat \\\hline
hard & low-level components &  VHDL designs \\\hline
\end{tabular}
\caption{\label{tab:softhard} Different VM implementation types, their requirements, and an example.}
\end{center}
\end{footnotesize}
\end{table}

There are many reasons why a semi-hard implementation was desired for Fhat and these reasons will be articulated in the sections discussing the various components of the Fhat architecture. However, while this section presents the semi-hard implementation, a soft implementation of Fhat called reduced Fhat (r-Fhat) will be briefly discussed. In short, r-Fhat is faster than the Fhat virtual machine, but does not support run-time machine portability and machine-level reflection. In other words, r-Fhat does not support those functions that require an RDF representation of the machine state.

\begin{figure}[h!]
	\centering
		\includegraphics[width=0.75\textwidth]{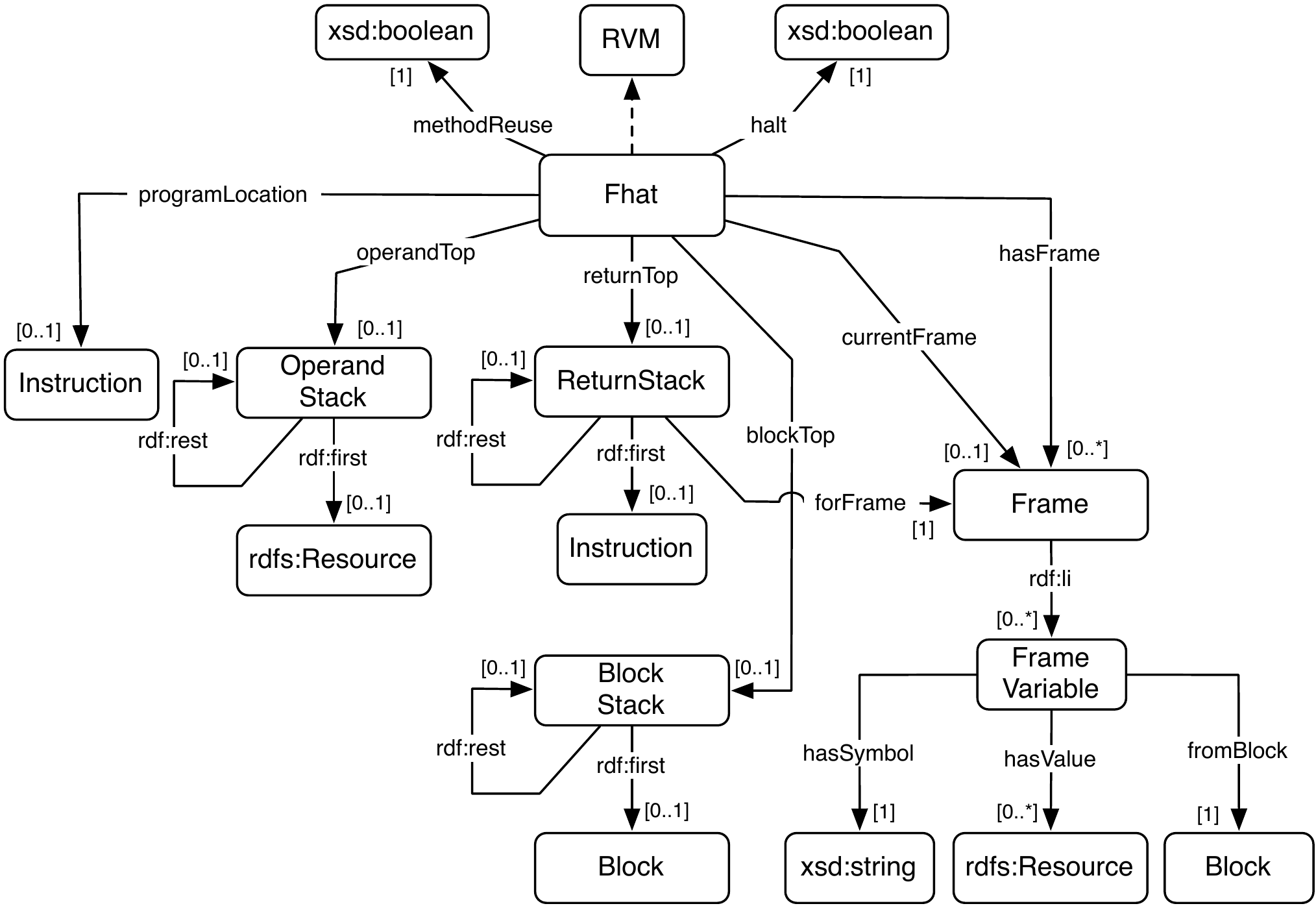}
	\caption{The ontological model of the Fhat virtual machine.}
	\label{fig:fhat}
\end{figure}

Any high-level language can be written to take advantage of the Fhat architecture. While Neno and Fhat were developed in concert and thus, are strongly connected in their requirements of one another, any language that compiles to Fhat RDF triple-code can use a Fhat RVM. This section will discuss the Fhat RVM before discussing the Fhat instruction set. Figure \ref{fig:fhat} presents the Fhat machine architecture. This machine architecture is represented in OWL and is co-located with other resources in the ontology layer of the RDF network.

There are $8$ primary components to the Fhat RVM. These are enumerated below for ease of reference. Each component will be discussed in more detail in the following subsections.

\begin{enumerate}
\item \ttt{Fhat}: the CPU that interprets instructions and uses its various components for processing those instructions.
\item \ttt{halt}: suspends Fhat processing when false, and permits processing when true.
\item \ttt{methodReuse}: determines whether or not method triple-code is reused amongst object instances.
\item \ttt{programLocation}: a pointer to the current instruction being executed (i.e.~a PC).
\item \ttt{BlockStack}: an \ttt{rdf:List} that can be pushed and popped for entering and exiting blocks.
\item \ttt{OperandStack}: an \ttt{rdf:List} that can be pushed and popped for arithmetic computations.
\item \ttt{Frame}: a \ttt{Method} unique environment for storing local variables.
\item \ttt{ReturnStack}: an \ttt{rdf:List} that provides a reference to the instruction that called a method and the frame of that method.
\end{enumerate}

While all of these components are represented in RDF, only \ttt{Fhat} has an external software component. The software implementation of \ttt{Fhat} is called the ``virtual machine process" in Figure \ref{fig:system-architecture}.\footnote{When the term Fhat is used, it is referring to the entire virtual machine, when the teletyped term \ttt{Fhat} is used, it is referring to the virtual machine process identified by the URI \ttt{Fhat}.}

\subsubsection{\ttt{Fhat}}

\ttt{Fhat} is the primary component of the Fhat RVM. \ttt{Fhat} is the most complicated component in the entire Fhat architecture. The high-level Neno pseudo-code for the \ttt{Fhat} component is
\begin{footnotesize}
\begin{verbatim}

Thing Fhat {
  execute() {
    while(!this.halt && this.programLocation != null) {
      Instruction i = this.programLocation
      if(i typeof Block) { ... }
      else if(i typeof If) { ... }
      else if(i typeof Expression) { ... }
      else if(i typeof Set) { ... }
      ...
      /* update programLocation */
    }
  }
}

\end{verbatim}
\end{footnotesize}
The above pseudo-code should be implemented in the language of the virtual machine process and thus, for the executing hardware CPU. 

It is worth noting that a \ttt{Fhat} virtual machine process can be written in Neno as demonstrated in the Neno code above. For example, assume a Neno implemented Fhat instance called Fhat1. In such cases, another Fhat, called Fhat2, is processing Fhat1.  Fhat2 can be run on yet another Fhat, called Fhat3, or grounded into some other language that is translating code to the native machine language. This is possible because Neno/Fhat is Turing complete and thus, can run a simulation of itself. When a simulation of itself is run, a complete RDF virtual machine is created. In this simulation environment, both the state and process of the Fhat RVM are represented in RDF.

The current version of Fhat supports most common uses of the \ttt{xsd:anySimpleType} and a few of these uses are summarized below:

\begin{itemize}
\item \ttt{xsd:boolean}: \ttt{Not}, \ttt{Equals}
\item \ttt{xsd:integer}, \ttt{xsd:float}, \ttt{xsd:double}: \ttt{Arithmetic}, \ttt{Compare}
\item \ttt{xsd:string}: \ttt{Add}, \ttt{Compare}
\item \ttt{xsd:date}, \ttt{xsd:dateTime}: \ttt{Add}, \ttt{Subtract}, \ttt{Compare}
\item \ttt{xsd:anyURI}: \ttt{Compare}.
\end{itemize}

\subsubsection{\ttt{halt}}

At any time, \ttt{Fhat} can be forced to halt by setting the \ttt{halt} property of \ttt{Fhat} to \ttt{true$^\wedge$$^\wedge$xsd:boolean}. Multi-threading can be simulated in this way. A Neno program can be engineered to run a master \ttt{Fhat} that has a reference to the \ttt{halt} property of all its slave \ttt{Fhat}s. By setting the \ttt{halt} property, the \ttt{Fhat} master can control which \ttt{Fhat} slaves are able to process at any one time. In essence, the master \ttt{Fhat} serves as an operating system.

\subsubsection{\ttt{methodReuse}}

When \ttt{methodReuse} is set to \ttt{true$^\wedge$$^\wedge$xsd:boolean}, \ttt{Fhat} will instantiate new objects with unique instructions for each method. When \ttt{methodReuse} is set to \ttt{false$^\wedge$$^\wedge$xsd:boolean}, \ttt{Fhat} will reuse method triple-code amongst the same methods for the different objects. This will be discussed in more detail in \S \ref{sec:methodreuse}.

\subsubsection{\ttt{programLocation}}

The \ttt{programLocation} is a pointer to the current instruction being executed by Fhat. Fhat executes one instruction at a time and thus, the \ttt{programLocation} must always point to a single instruction. The ``while" loop of Fhat simply moves the \ttt{programLocation} from one instruction to the next. At each instruction, Fhat interprets what the instruction is (by its \ttt{rdf:type} ``opcode") and uses its various components appropriately. When there are no more instructions (i.e.~when there no longer exists a \ttt{programLocation} property), Fhat halts. 

\subsubsection{\ttt{BlockStack}}

The \ttt{BlockStack} is important for variable setting. When a new variable is created in a block of code, it is necessary to associate that variable with that block. When the thread of execution exits the block, all variables created in that block are dereferenced (i.e.~deallocated).

\subsubsection{\ttt{OperandStack}}

The \ttt{OperandStack} is a LIFO (i.e.~``last in, first out") stack that supports any \ttt{rdfs:Resource}. The \ttt{OperandStack} is used for local computations such as \ttt{x = 1 + (2 * 3)}. For example, when \ttt{x = 1 + (2 * 3)} is executed by \ttt{Fhat}, \ttt{Fhat} will

\begin{enumerate}
\item push the value 1 on the \ttt{OperandStack}
\item push the value 2 on the \ttt{OperandStack}
\item push the value 3 on the \ttt{OperandStack}
\item pop both 2 and 3 off the \ttt{OperandStack}, multiply the two operands, and push the value 6 on the \ttt{OperandStack}
\item pop both 1 and 6 the \ttt{OperandStack}, add the two operands, and push the value 7 on the \ttt{OperandStack}
\item set the current \ttt{Frame} \ttt{FrameVariable} \ttt{x} to the value $7$ popped off the \ttt{OperandStack}.
\end{enumerate}

\begin{figure}[h!]
	\centering
		\includegraphics[width=0.5\textwidth]{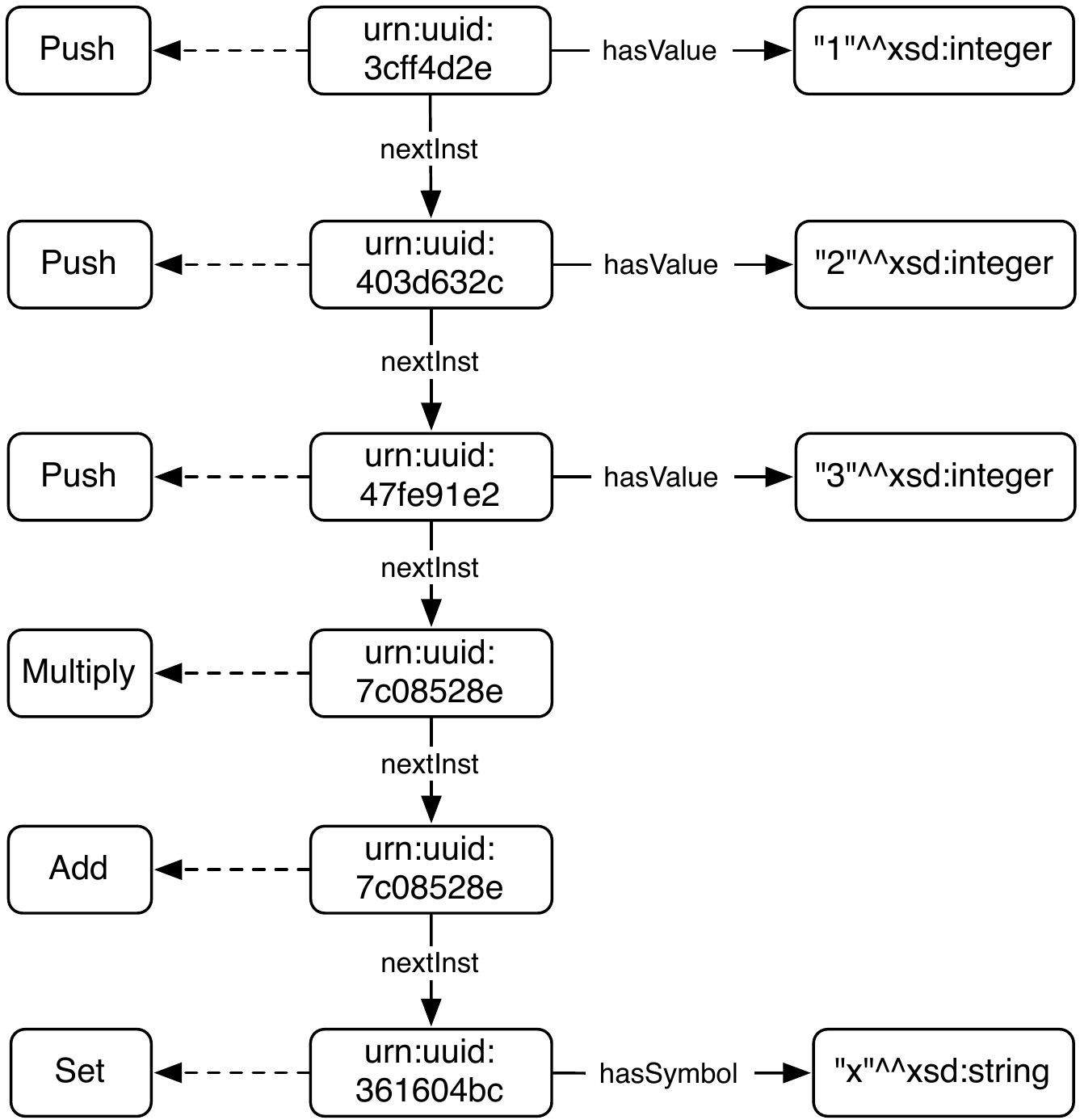}
	\caption{The triple-code representation of the statement \ttt{x = 1 + (2 * 3)}.}
	\label{fig:example-add-mult}
\end{figure}

The Neno statement \ttt{x = 1 + (2 * 3)} is actually multiple instructions when compiled to Fhat triple-code. The NenoFhat compiler would translate the statement to the triple-code represented in Figure \ref{fig:example-add-mult}.

It is very important to represent such components as the \ttt{OperandStack} component in RDF and not simply in the memory of the host CPU. Suppose that a Fhat instance is to move to another physical machine or, by chance, lose its process ``back-end". If any of these two scenarios were the case, the state of the machine is always saved in RDF and thus, would simply ``freeze" to await another virtual machine process to continue its execution. If the \ttt{OperandStack} was represented in software and thus, in RAM, then when the software halted, the \ttt{OperandStack} would be lost and the state of the machine would be inconsistent with its \ttt{programLocation}. With an RDF state, the RAM representation of the virtual machine process has a negligible effect on the consistency of the machine.

\subsubsection{\ttt{Frame}}

Fhat is a frame-based processor. This means that each invoked method is provided a \ttt{Frame}, or local environment, for its variables (i.e.~\ttt{FrameVariable}s). Due to how variables are scoped in object-oriented languages and because Neno does not support global variables, each method can only communicate with one another through parameter (i.e.~method arguments) passing, return value passing, or object manipulations. When method $A$ calls method $B$, the parameters passed by method $A$ are stored in method $B$'s \ttt{Frame} according to the variable names in the method description. For example, assume the following method,
\begin{footnotesize}
\begin{verbatim}

xsd:integer methodB(xsd:integer a) {
  return a + "1"^^xsd:integer;
}

\end{verbatim}
\end{footnotesize}
If method $A$ calls method $B$, with the statement,
\begin{footnotesize}
\begin{verbatim}

xsd:integer x = marko.methodB("2"^^xsd:integer);

\end{verbatim}
\end{footnotesize}
the value $2$ is placed into the \ttt{Frame} of method $B$ with the associated variable \ttt{a}. Method $B$ adds $1$ to the value and pushes the value $3$ on the \ttt{OperandStack}. Method $A$ pops one value off the \ttt{OperandStack} and sets the local variable \ttt{x} to the value $3$. The \ttt{OperandStack} is used for the placement of method return values.

\subsubsection{\ttt{ReturnStack}}

The \ttt{ReturnStack} is a LIFO stack that maintains pointers to the return location of a method and the method \ttt{Frame}. To support recursion, the \ttt{ReturnStack} maintains a pointer to the specific \ttt{Frame} that is being returned to.

In order to explain how the \ttt{ReturnStack} is used, an example is provided. When method $A$ calls method $B$, the next instruction of method $A$ following the method invocation instruction is pushed onto the \ttt{ReturnStack}. When method $B$ has completed its execution (e.g.~a \ttt{return} is called), Fhat pops the instruction off the \ttt{ReturnStack} and sets its \ttt{programLocation} to that instruction. In this way, control is returned to method $A$ to complete its execution. When \ttt{return} is called in method $B$, Fhat will delete (i.e.~deallocate) all triples associated with the method $B$ \ttt{Frame}. If \ttt{return} has a value (e.g. \ttt{return 2}), that value is pushed onto the \ttt{OperandStack} for method $A$ to use in its computation.

\subsection{Migrating Fhat Across Different Host CPUs\label{sec:migrate}}

\begin{figure}[h!]
	\centering
		\includegraphics[width=0.4\textwidth]{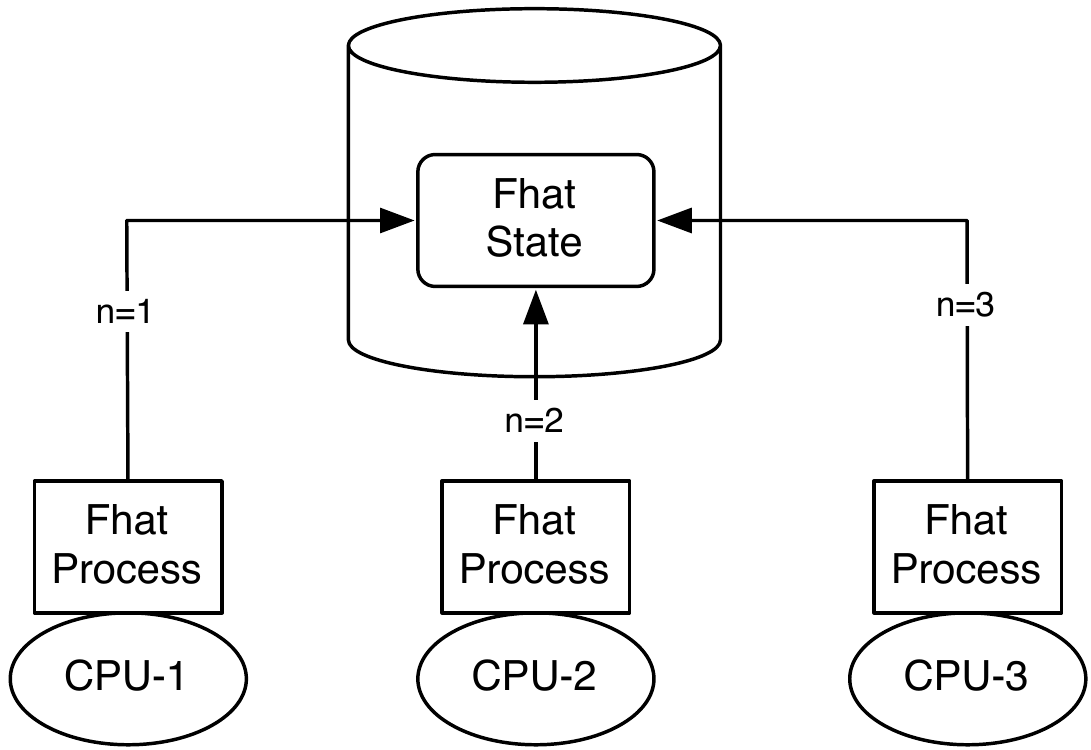}
	\caption{Migrating a Fhat execution across multiple host CPUs.}
	\label{fig:migrate-fhat}
\end{figure}

An interesting aspect of Fhat is the ability to migrate a Fhat process across various host CPUs. A Fhat implementation has two primary components: an RDF state representation and a software process. Because both the RDF triple-code and the complete state of a Fhat instance is represented in the RDF network, it does not matter which Fhat process is executing a particular Fhat state. The Fhat RDF state representation ensures that there are no global variables in the software process. The only variables created in the software process are local to the instruction being executed. Because there are no global variables in the software process, any software process can execute the Fhat RDF state without requiring inter-software process communication. For example, one host CPU  can be running the Fhat software process and halt. Another CPU can then start another Fhat software process that points to the URI of the originally halted Fhat RDF state and continue its execution. This concept is diagrammed in Figure \ref{fig:migrate-fhat} where $n=1$ refers to instruction 1.

\begin{figure}[h!]
	\centering
		\includegraphics[width=0.5\textwidth]{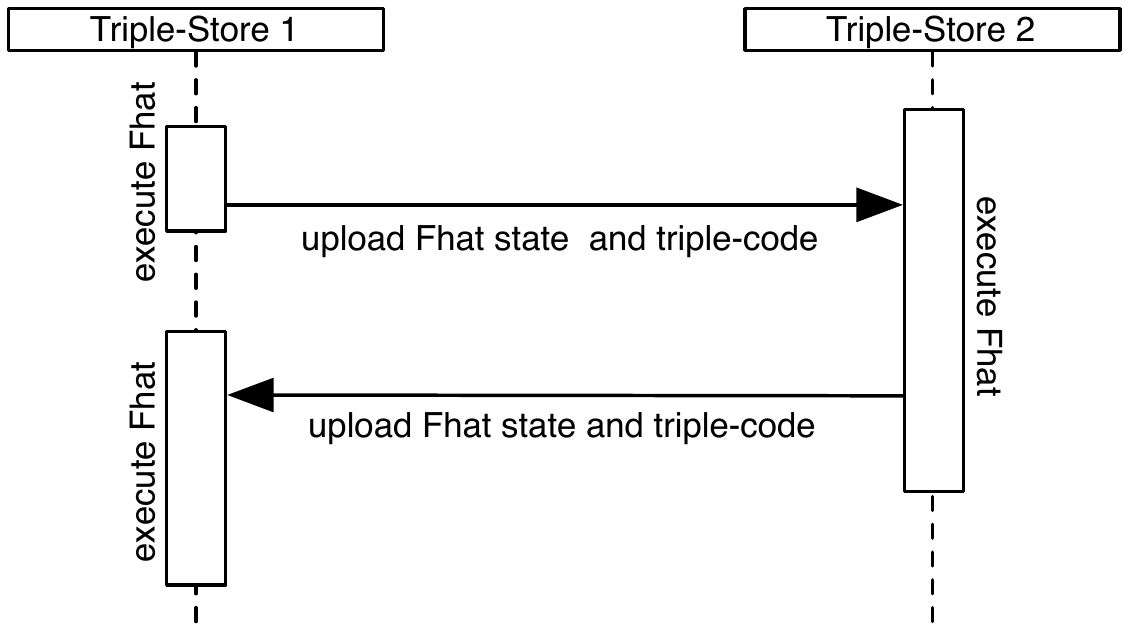}
	\caption{Migrating a Fhat state across different triple-stores.}
	\label{fig:upload-fhat}
\end{figure}

In principle, each CPU can execute one instruction and then halt. In this way, it is possible to migrate the Fhat RVM across different host CPU's. Thus, if a portion of the Semantic Web is needed for a particular computation, it may be best to have the physical computer supporting that RDF sub-network host the Fhat RVM. Once the Fhat RVM has completed computing that particular RDF sub-network, it can halt and another CPU can pick up the process on a yet another area of the Semantic Web that needs computing by the Fhat RVM. In this model of computing, data doesn't move to the process, the process moves to the data. This idea is diagrammed in Figure \ref{fig:upload-fhat}, where both triple-store servers have Fhat process implementations.

\subsection{Fhat Reflection\label{sec:reflect}}

\begin{figure}[h!]
	\centering
		\includegraphics[width=0.6\textwidth]{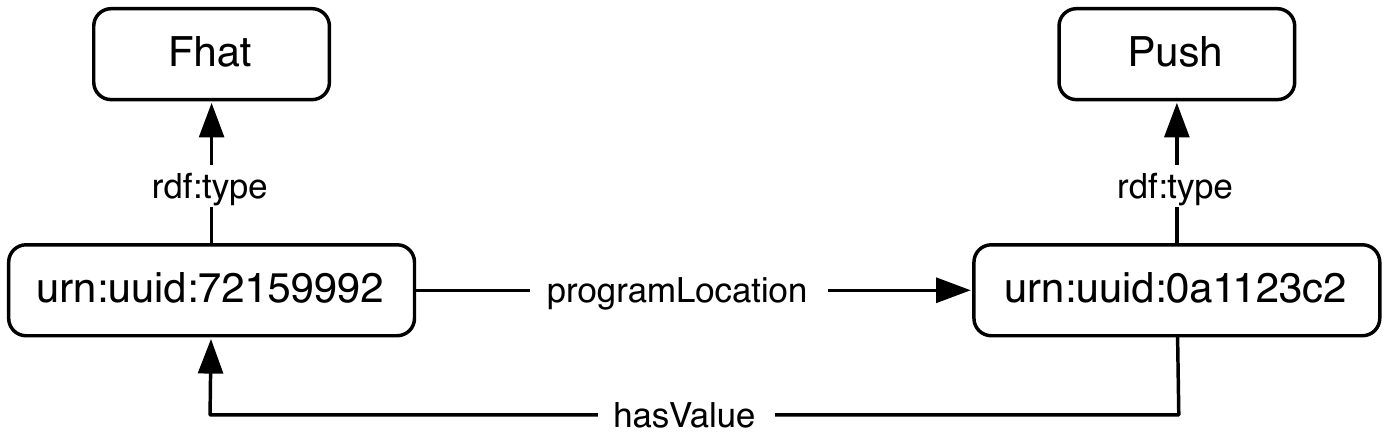}
	\caption{A Fhat processor can process itself.}
	\label{fig:example-self-reflective}
\end{figure}

A Fhat RVM and the triple-code that it is executing are in the same address space and thus, can reference one another. It is the UUID address space of Neno/Fhat that makes it a unique programming environment in that Neno is not only a completely reflective language, but also that it removes the representational stack found in most other programming environments. Language reflection means that the program can modify itself during its execution. Many scripting languages and even Java (through the \ttt{java.lang.reflect} package) support language reflection. However, not only does Neno/Fhat support language reflection, it also supports machine reflection. A Fhat can modify itself during its execution. There are no true boundaries between the various components of the computation. This idea is represented in Figure \ref{fig:example-self-reflective}, where a Fhat RVM has its program counter (\ttt{programLocation}) pointing to a \ttt{Push} instruction. The \ttt{Push} instruction is instructing Fhat to push a reference to itself on its operand stack. With a reference to the Fhat instance in the Fhat operand stack, Fhat can manipulate its own components. Thus, the Fhat RVM is executing triple-code that is manipulating itself.

\subsection{r-Fhat\label{sec:rfhat}}

What has been presented thus far is a semi-hard implementation of Fhat. The semi-hard implementation explicitly encodes the state of a Fhat instance in RDF. While this has benefits such as fault tolerance due to virtual machine process failures, support for distributed computing in the form of processor migration, and support for machine-based evolutionary algorithms, it requires a large read/write overhead. Each instruction requires the virtual machine process to explicitly update the virtual machine state.  A faster Fhat virtual machine can be engineered that does not explicitly encode the state of the machine in the RDF network. In such cases, the only read/write operations that occur are when an object is instantiated, destroyed, or a property manipulated. This faster Fhat is called reduced Fhat (r-Fhat). In r-Fhat, the operand stack, return stack, etc. are data structures in the implementing language. r-Fhat does not have an OWL machine architecture nor an RDF state.

\section{The Fhat Instruction Set}

In order for Neno software to run on a Fhat machine instance, it must be compiled to a Fhat OWL API that is compliant with the Fhat instruction set (the Fhat OWL API \ttt{owl:imports} the Fhat instruction set ontology). A Fhat RVM uses the Fhat OWL API as a ``blueprint" for constructing the instance-level representation of the RDF triple-code. It is the instance-level triple-code that the Fhat RVM ``walks" when a program is executing.

\subsection{The \ttt{Method}}

\begin{figure}[h!]
	\centering
		\includegraphics[width=0.6\textwidth]{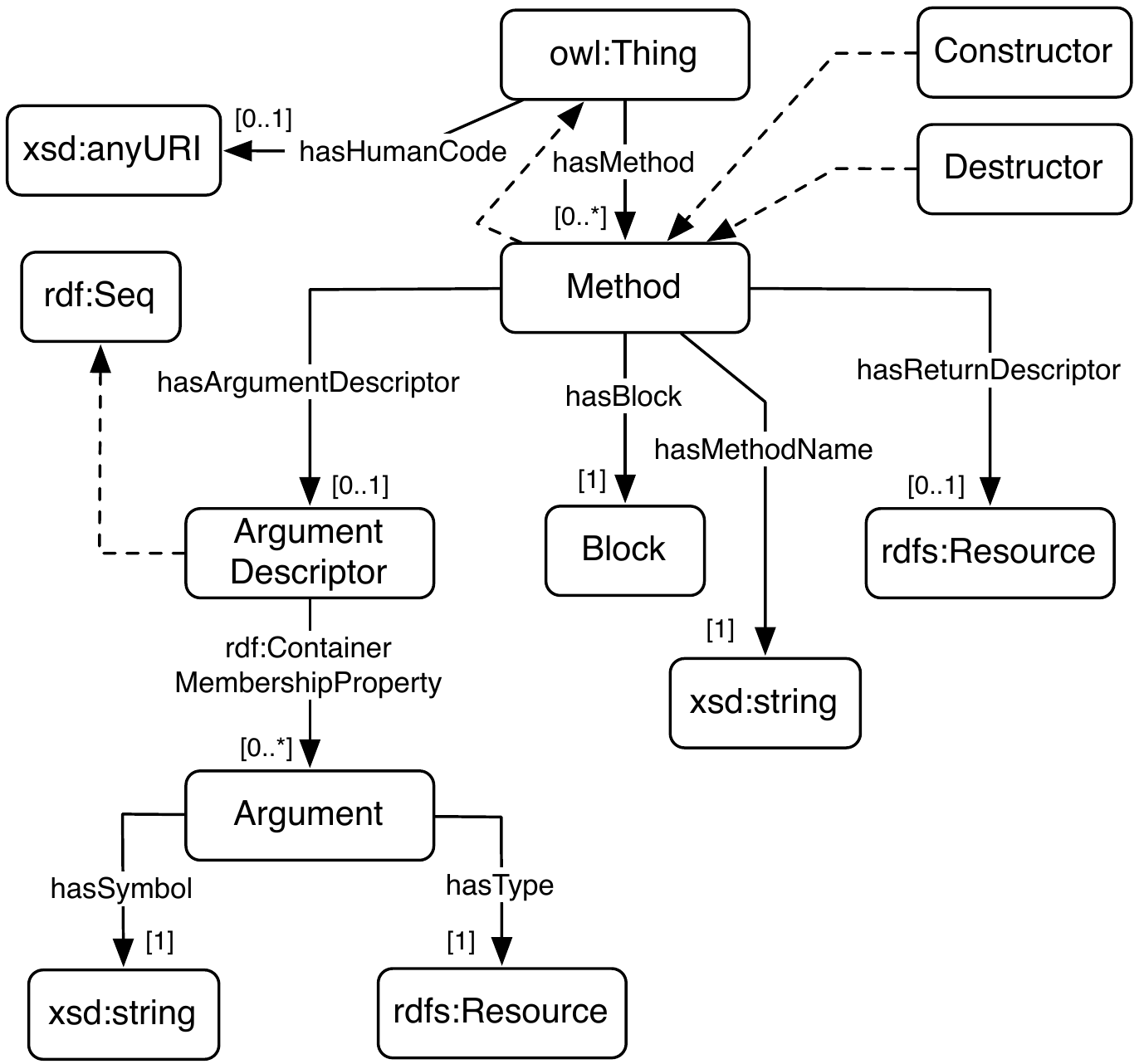}
	\caption{The OWL \ttt{Method} ontology.}
	\label{fig:method}
\end{figure}

In Neno, the only process code that exists is that which is in a \ttt{Method} \ttt{Block}. Figure \ref{fig:method} defines the OWL ontology of a \ttt{Method}.

A \ttt{Method} has an \ttt{ArgumentDescriptor} that is of \ttt{rdfs:subClassOf} \ttt{rdf:Seq} and a return descriptor that is of type \ttt{rdfs:Resource}. The sequence of the \ttt{ArgumentDescriptor} \ttt{Argument} denotes the placement of the \ttt{Method} parameter in the method declaration. For instance,
\begin{footnotesize}
\begin{verbatim}

xsd:integer exampleMethod(xsd:string n, Human h) { ... }

\end{verbatim}
\end{footnotesize}
would set the object of the \ttt{hasReturnDescriptor} property to the URI \ttt{xsd:integer} and the \ttt{ArgumentDescriptor} to the \ttt{Argument}s \ttt{n} (\ttt{rdf:\_1}) and \ttt{h} (\ttt{rdf:\_2}). 

The \ttt{hasHumanCode} property can be used, if desired, to point to the original human readable/writeable source code that describes that class and its methods. By using the \ttt{hasHumanCode} property, it is possible for ``in-network" or run-time compiling of source code. In principle, a Neno compiler can be written in Neno and be executed by a Fhat RVM. The Neno compiler can compile the representation that results from resolving the URI that is the value of the \ttt{xsd:anyURI}.

\subsubsection{A \ttt{Block} of Fhat Triple-Code}

\begin{figure}[h!]
	\centering
		\includegraphics[width=0.6\textwidth]{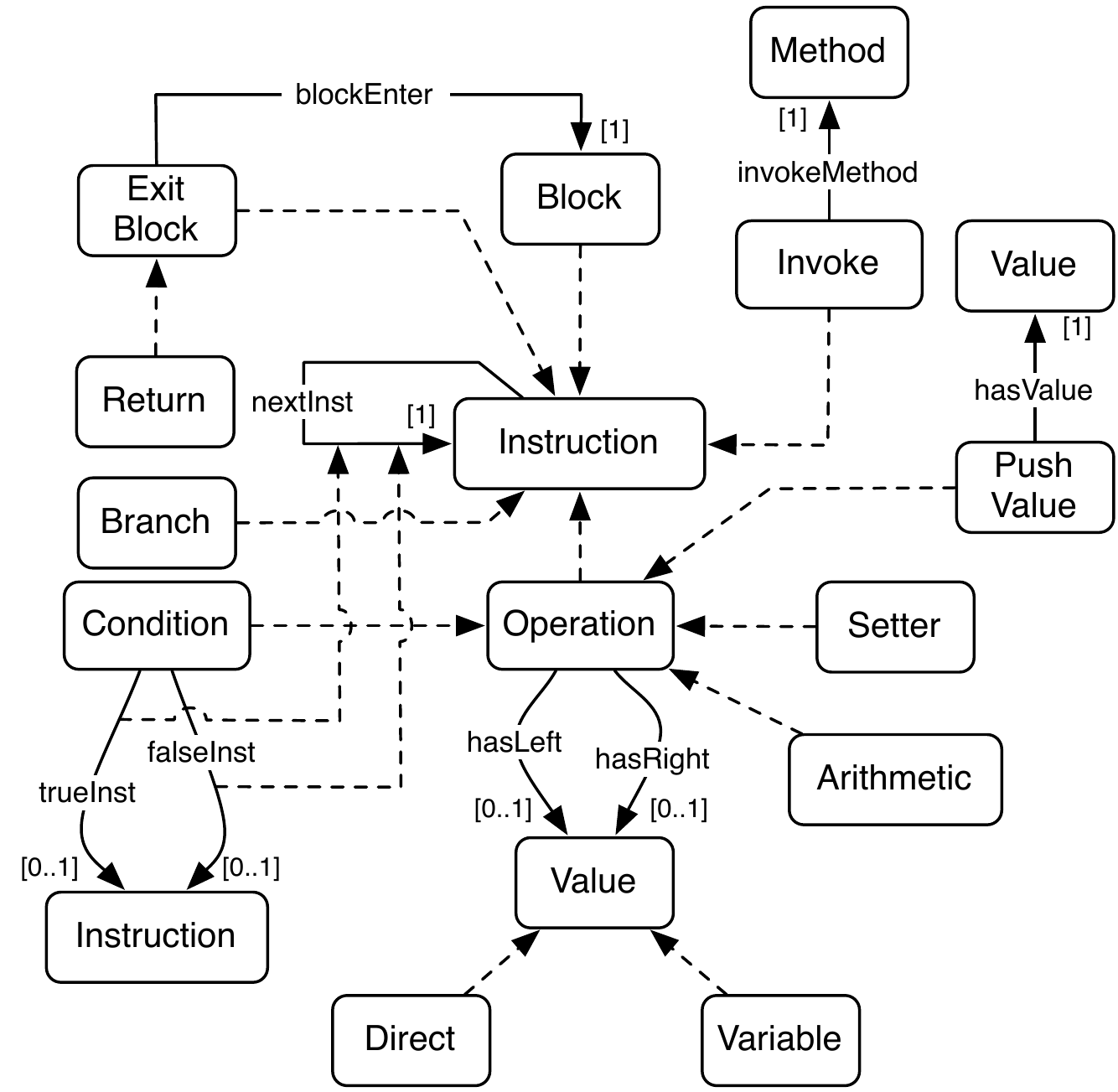}
	\caption{The OWL ontology for a \ttt{Block} of \ttt{Instruction}s.}
	\label{fig:block}
\end{figure}

A \ttt{Method} has a single \ttt{Block}. A \ttt{Block} is an \ttt{rdfs:subClassOf} \ttt{Instruction} and is composed of a sequence of \ttt{Instruction}s. The \ttt{Instruction} sequence is denoted by the \ttt{nextInst} property. The \ttt{Instruction} \ttt{rdf:type} is the ``opcode" of the \ttt{Instruction}. The set of all \ttt{Instruction}s is the instruction set of the Fhat architecture. Figure \ref{fig:block} provides a collection of the super class \ttt{Instruction}s that can exist in a \ttt{Block} of code and their relationship to one another.

Examples of these super classes are itemized below.\footnote{\ttt{Condition}s are unique in that they have a \ttt{trueInst} and a \ttt{falseInst} property. If the \ttt{Condition} is true, the next \ttt{Instruction} is the one pointed to by the \ttt{trueInst} property, else the next \ttt{Instruction} is one pointed to by the \ttt{falseInst} property.}
\begin{itemize}
	\item \ttt{Arithmetic}: \ttt{Add}, \ttt{Divide}, \ttt{Multiply}, \ttt{Not}, \ttt{Subtract}.
	\item \ttt{Condition}: \ttt{Equals}, \ttt{GreaterThan}, \ttt{GreaterThanEqual}, \ttt{LessThan}, \ttt{LessThanEqual}.
	\item \ttt{Setter}: \ttt{NetQuery}, \ttt{Set}, \ttt{SetClear}, \ttt{SetMinus}, \ttt{SetPlus}, \ttt{SetQuery}.
	\item \ttt{Invoke}: \ttt{Construct}, \ttt{Destruct}.
\end{itemize}

The \ttt{Value} class has a set of subclasses. These subclasses are itemized below.
\begin{itemize}
	\item \ttt{Direct}: \ttt{LocalDirect}, \ttt{PopDirect}.
	\item \ttt{Variable}: \ttt{LocalVariable}, \ttt{FieldVariable}, \ttt{ObjectVariable}.
\end{itemize}

When a Fhat instance enters a \ttt{Method} it creates a new \ttt{Frame}. When a \ttt{Variable} is declared, that \ttt{Variable} is specified in the \ttt{Frame} and according to the current \ttt{Block} of the \ttt{Fhat} instance as denoted by \ttt{Fhat}'s \ttt{blockTop} property. A \ttt{Block} is used for variable scoping. When Fhat leaves a \ttt{Block}, it destroys all the \ttt{FrameVariable}s in the current \ttt{Frame} that have that \ttt{Block} as their \ttt{fromBlock} property (refer to Figure \ref{fig:fhat}). However, entering a new \ttt{Block} is not exiting the old \ttt{Block}. Parent \ttt{Block} \ttt{FrameVariable}s can be accessed by child \ttt{Block}s. For instance, in the following Neno code fragment,
\begin{footnotesize}
\begin{verbatim}

xsd:integer x = "1"^^xsd:integer;
if(x > 2) {
  xsd:integer y = x;
}
else{
  xsd:integer y = x;
}

\end{verbatim}
\end{footnotesize}
the two \ttt{y} \ttt{Variable}s in the if and else \ttt{Block}s are two different \ttt{FrameVariables} since they are from different \ttt{Blocks}. Furthermore, note that both the if and else \ttt{Block}s can access the value of \ttt{x} since they are in the child \ttt{Block} of the \ttt{Block} declaring the variable \ttt{x}. When Fhat leaves a \ttt{Method} (i.e.~returns), its \ttt{Frame} and its \ttt{FrameVariable}s are destroyed through dereferencing.

\subsection{A \ttt{Method} Instance\label{sec:methodreuse}}

\begin{figure}[h!]
	\centering
		\includegraphics[width=0.6\textwidth]{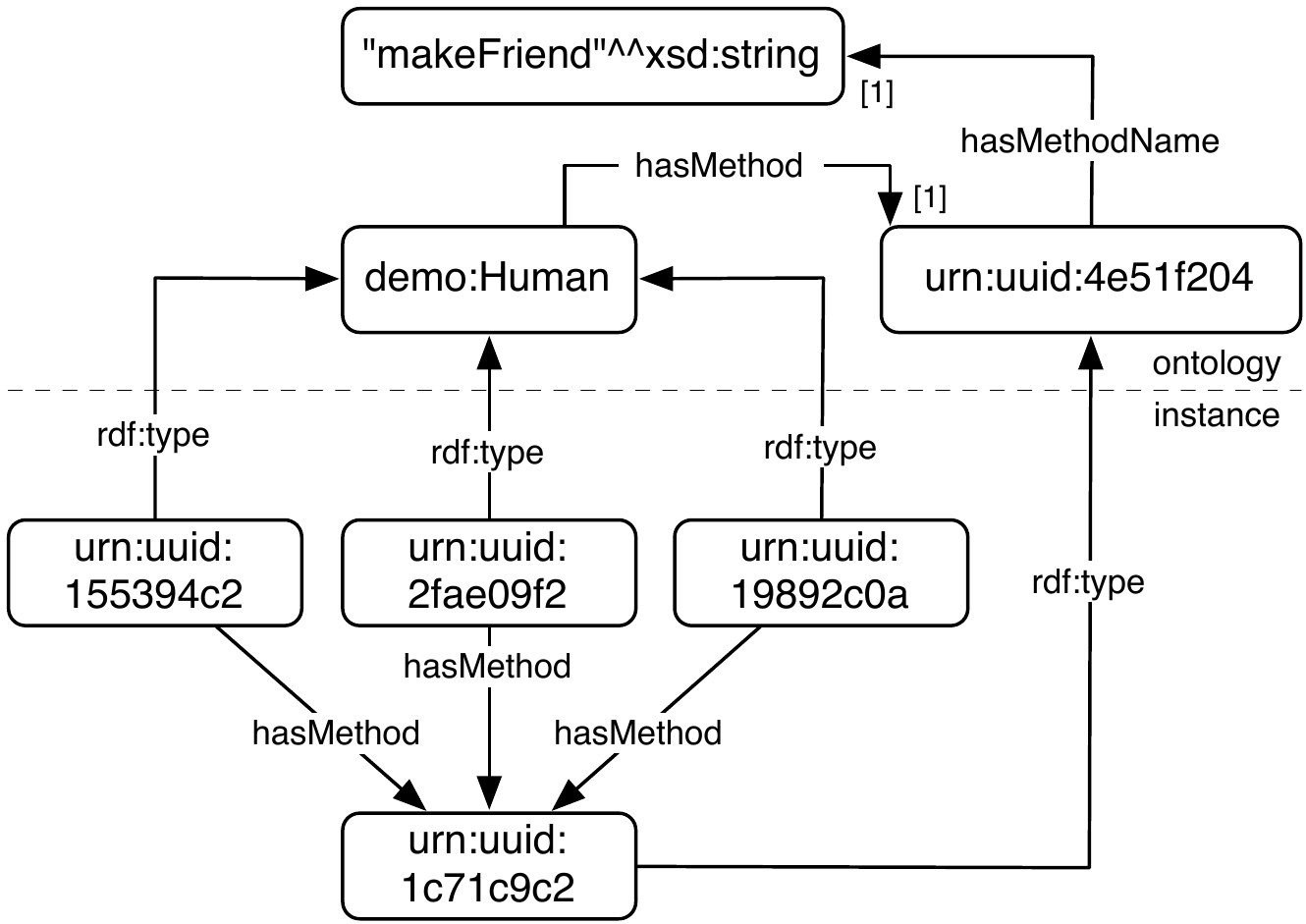}
	\caption{Multiple object's of the same type will share the same \ttt{Method} instance.}
	\label{fig:example-shared-method}
\end{figure}

There are two ways in which a \ttt{Method} instance is handled by Fhat: global and local instance models. In the global instance model, when a new object is instantiated, its methods are also instantiated. However, if the instantiated \ttt{Method} already exists in the RDF network, the newly created object points its \ttt{hasMethod} property to a previously created \ttt{Method} of the same \ttt{hasMethodName} and UUID. Thus, only one instance of a \ttt{Method} exists for all the objects of the same class type. While it is possible to have a unique \ttt{Method} instance for each object, by supporting method reuse amongst objects, Fhat limits the growth (in terms of the number of triples) in the RDF network. Furthermore, this increases the speed of the Fhat RVM since it does not need to create a new \ttt{Method} from the Fhat OWL API of that \ttt{Method}. The global instance model is diagrammed in Figure \ref{fig:example-shared-method}. To ensure global instances, the \ttt{methodReuse} property of the \ttt{Fhat} instance is set to \ttt{"true"$^\wedge$$^\wedge$xsd:boolean} (refer to Figure \ref{fig:fhat}).

\begin{figure}[h!]
	\centering
		\includegraphics[width=0.6\textwidth]{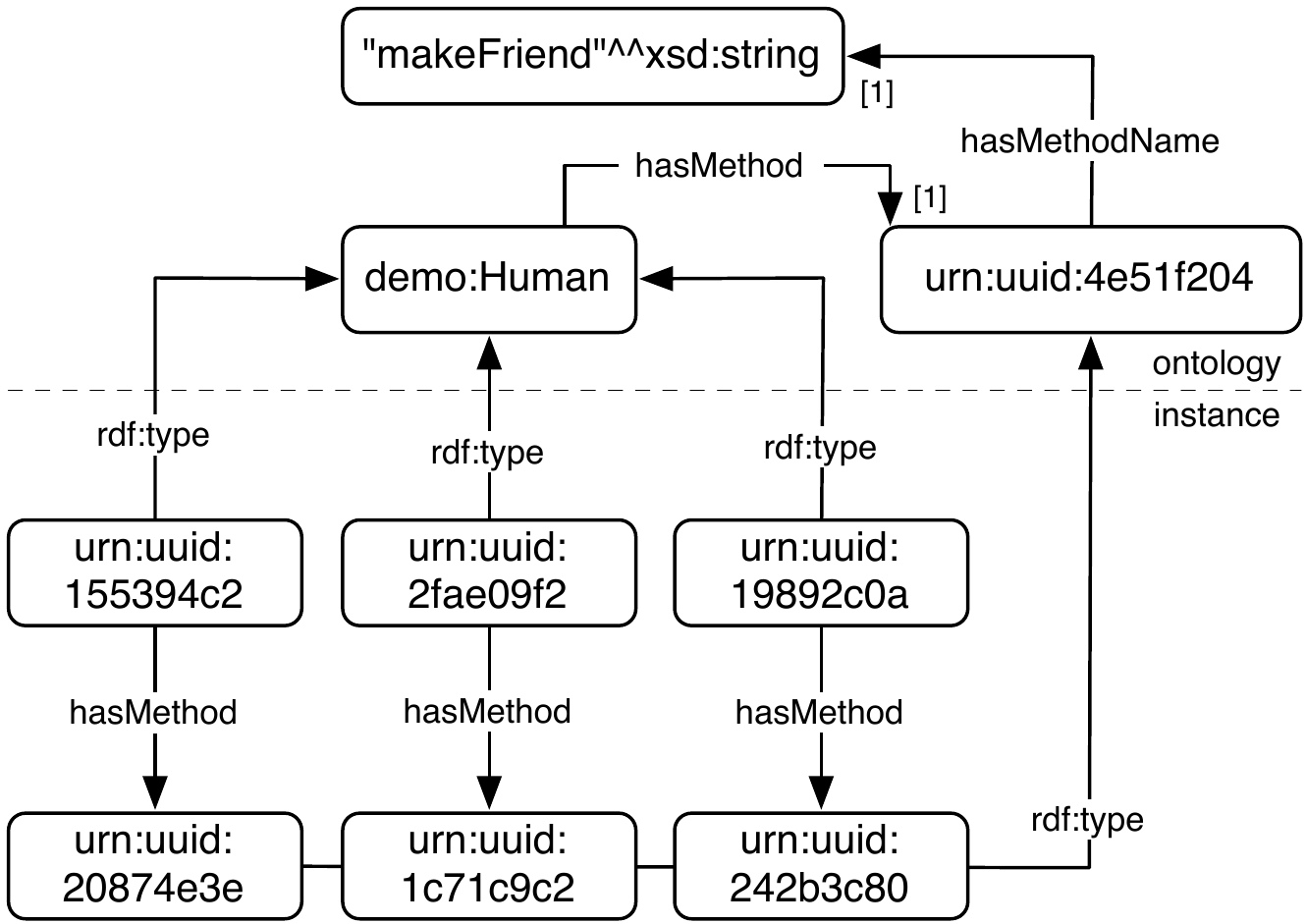}
	\caption{Multiple object's of the same type each have a unique \ttt{Method} instance.}
	\label{fig:example-unique-method}
\end{figure}

In the local instance model, the \ttt{methodReuse} property of a \ttt{Fhat} instance is set to \ttt{"false"$^\wedge$$^\wedge$xsd:boolean}. In such cases, a new \ttt{Method} instance is created with each new instance of an \ttt{owl:Thing}. The benefit of this model is that method reflection can occur on a per-object basis. If an object is to manipulate its \ttt{Method} triple-code at run-time, it can do so without destroying the operation of its fellow \ttt{owl:Thing}s. The drawback of the local instance model is triple-store ``bloat" and an increase in the time required to instantiate an object relative to the global instance model. The local instance model is diagrammed in Figure \ref{fig:example-unique-method}.

\subsubsection{An Example \ttt{Method} Instance}

Suppose the following code,
\begin{footnotesize}
\begin{verbatim}

owl:Thing demo:Human
{
  xsd:int example(xsd:string a) {
    if(a == "marko"^^xsd:string) {
      return "1"^^xsd:int;
    }
    else {
      return "2"^^xsd:int;
    }
  }
}

\end{verbatim}
\end{footnotesize}

\begin{figure}[h!]
	\centering
		\includegraphics[width=0.75\textwidth]{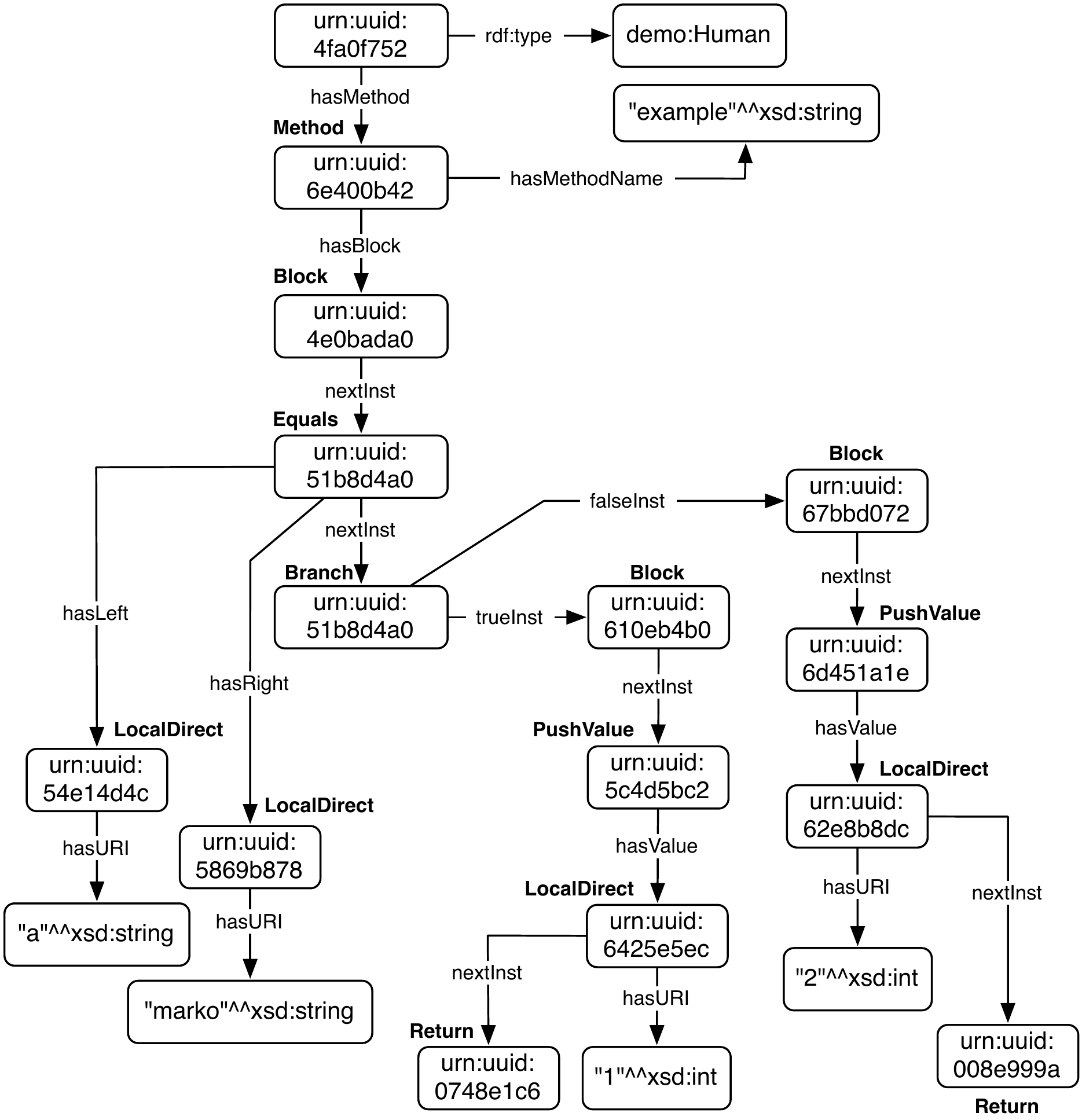}
	\caption{The RDF triple-code for the \ttt{example(xsd:string)} method.}
	\label{fig:example-triple-code}
\end{figure}

When this code is compiled, it compiles to a Fhat OWL API. When an instance of \ttt{demo:Human} is created, the Fhat RVM will start its journey at the URI \ttt{demo:Human} and move through the ontology creating instance UUID URIs for all the components of the \ttt{demo:Human} class. This includes, amongst its hard-coded properties, its \ttt{Method}s, their \ttt{Block}s, and their \ttt{Instruction}s. When the \ttt{demo:Human} class is instantiated, an instance will appear in the RDF network as diagrammed in Figure \ref{fig:example-triple-code}.

\section{Conclusion}

The primary drawback of the RVM is computing time because a virtual machine must not only read the RDF network to interpret the program instructions, it must also read/write to the RDF network to manipulate data (i.e.~instance objects) in the RDF network. Moreover, the virtual machine must read and write to that sub-network of the RDF network that represents the virtual machine's state (e.g.~program counter, operand stack, etc.). In doing so, many read/write operations occur in order for the virtual machine to compute. However, much of this issue can be resolved through the use of r-Fhat.

Imagine a world where virtual machines are as easy to distribute as an HTML document (e.g.~an RDF/XML encoding of the virtual machine sub-network). Given that a virtualized machine encodes its state in the RDF network, think about how RVMs can ``move" between physical machines in mid-execution. There is complete hardware independence as no physical machine maintains a state representation. Physical machines compute the RVM by reading its program location, its operand stack, its heap, etc. and update those data structures. In such situations, a personal computer can be encoded in the RDF network and be accessed anywhere. Thus, the underlying physical machine is only a hardware shell for the more ``personal" machine encoded in the RDF network. These ideas are analogous to those presented in \cite{pervasive:satya2007}.

In the RDF network, RVMs, APIs, and triple-code are ``first-class" web entities. What happens when archiving services such as the Internet Archive, search engine caches, and digital libraries archive such RDF programs and ``snap-shot" states of the executing RVMs \cite{preserve:lorie2001,wi:nelson2007}? In theory, the state of computing world-wide, can be saved/archived and later retrieved to resume execution. The issues and novelties that archiving computations presents are many and are left to future work in this area.

Much of the Semantic Web effort is involved in the distribution of knowledge between organizational boundaries \cite{sem:davies2003}. This is perhaps the primary purpose of the ontology. In this respect, organizations of a similar domain should utilize shared ontologies in order to make their information useable between their respective organizations. Procedural encodings support the distribution of not only the knowledge models, but also the algorithms that can be applied to compute on those models. In a non-disjoint manner, data and code are easily exchanged between organizational boundaries \cite{usingrdf:rodriguez2007}.

Given that the RDF network is composed of triples and triples are composed of URIs and literals, the address space of any virtual machine in the RDF network is the set of all URIs and literals. Given that there are no bounds to the size of these resources, there are no realistic space limitations on the RVM.  In other words, the amount of disk-space provided world-wide to support the Semantic Web is the actual memory constraints of this model. However, the success of this distributed computing paradigm relies on the consistent use of such standards as the Link Data specification \cite{berners:ldata2006}. With further developments in Linked Data models and the RVM model of computing, the Semantic Web can be made to behave like a general-purpose computer.

\section*{Acknowledgments}

This research was made possible by a generous grant from the Andrew W. Mellon Foundation. Herbert Van de Sompel, Ryan Chute, and Johan Bollen all provided much insight during the development of these ideas.

        \bibliographystyle{apacite}
        \bibliography{../marko}

\end{document}